\documentclass[10pt,twocolumn,letterpaper]{article}

\usepackage{iccv}
\usepackage{times}
\usepackage{epsfig}
\usepackage{graphicx}
\usepackage{amsmath}
\usepackage{amssymb}
\usepackage{caption}  
\usepackage{times}
\usepackage{epsfig}
\usepackage{graphicx}
\usepackage{amsfonts}
\usepackage{amsmath}
\usepackage{subcaption}
\usepackage{amssymb}
\usepackage{mathtools}
\usepackage{booktabs}
\usepackage{algorithm}
\usepackage{caption,subcaption}
\usepackage{algorithmic}
\usepackage{mathrsfs} 
\usepackage{soul}
\usepackage{color}
\usepackage{multirow}
\usepackage{xcolor,colortbl}
\usepackage{soul} 
\usepackage{lipsum}
\usepackage{newtxtext,newtxmath}
\usepackage{comment}
\usepackage{titletoc}

\usepackage[numbers,sort]{natbib}

\definecolor{mistyrose}{rgb}{1.0, 0.89, 0.88}
\definecolor{palerobineggblue}{rgb}{0.59, 0.87, 0.82}
\definecolor{lavenderblue}{rgb}{0.9, 0.9, 1.0}
\definecolor{amagenta}{rgb}{1.0, 0.0, 0.88}
\definecolor{ablue}{rgb}{0.59, 0.5, 1.}
\definecolor{pptblue}{HTML}{00B0F0}
\definecolor{pptpink}{HTML}{E036CE}
\definecolor{pptred}{HTML}{FF3400}
\definecolor{pptred}{HTML}{FF3400}
\definecolor{pptblue}{HTML}{00B0F0}
\definecolor{pptpink}{HTML}{E036CE}

\newcommand{\datasetname}{{\emph{Video Person-Clustering Dataset}}\xspace}
\newcommand{\dnameemphasis}{{\emph{VPCD}}\xspace}
\newcommand{\methodnameMM}{{\emph{Multi-Modal High-Precision Clustering}}\xspace}
\newcommand{\mnameMM}{{\emph{MuHPC}}\xspace}

\newcommand{\new}[1]{\textcolor{black}{#1}}

\usepackage[pagebackref=true,breaklinks=true,colorlinks,bookmarks=false]{hyperref}
\newcommand{\todo}[1]{\textcolor{red}{[#1]}}

\iccvfinalcopy 

\def\iccvPaperID{1352} 

\title{Face, Body, Voice: Video Person-Clustering with Multiple Modalities}

\author{Andrew Brown$^{1}$, Vicky Kalogeiton$^{1,2}$, and Andrew Zisserman$^{1}$\\
$^{1}$VGG, Dept.\ of Engineering Science, University of Oxford, $^{2}$LIX, École Polytechnique, CNRS, IP Paris \\
{\tt\small \{abrown, az\}@robots.ox.ac.uk, vicky.kalogeiton@lix.polytechnique.fr}  \\
\small{\url{https://www.robots.ox.ac.uk/~vgg/data/Video_Person_Clustering/}}
}

\begin{document}
\maketitle

\begin{abstract}
The objective of this work is person-clustering in videos -- grouping characters according to their identity.
Previous methods focus on the narrower task of face-clustering, and for the most part ignore other cues such as the 
person's voice, their overall appearance (hair, clothes, posture), and
the editing structure of the videos. Similarly, most current datasets
evaluate only the task of face-clustering, rather than person-clustering.
This limits their applicability to downstream applications such as story understanding which require person-level, rather than only face-level,  reasoning.

In this paper we make contributions to address both these deficiencies:
first, we introduce a {\em Multi-Modal High-Precision Clustering algorithm} for
person-clustering in videos using cues from several modalities (face, body, and voice).
Second, we introduce a {\em Video Person-Clustering dataset}, for evaluating multi-modal person-clustering. 
It contains body-tracks for each annotated character, face-tracks when visible, and voice-tracks when speaking, with their associated features.
The dataset is by far the largest of its kind, and covers films and TV-shows representing a wide range of demographics. 
Finally, we show
the effectiveness of using multiple modalities for person-clustering, explore the use of this new broad task
for story understanding \new{through character co-occurrences}, and achieve a  new state of the art on all available datasets for face and person-clustering. 
\end{abstract}

\section{Introduction}
\label{sec:intro}

Clustering people by identity in videos is an appealing and  much-visited topic in
computer vision~\cite{cinbis2011iccv,Fitzgibbon02,tapaswi2014total,wu2013cvpr,jin2017iccv,Kalogeiton20,ball}.
It has several real-world applications, such as enabling 
person-specific browsing, organisation of video collections, character based fast-forwards, automatic cast listing; and story understanding, all without requiring any explicit identity labeling.  A
successful person-clustering framework can therefore alleviate the tremendous
annotation cost that is otherwise necessary for such applications.

\begin{figure}[t]
\begin{center}
\includegraphics[width=\linewidth]{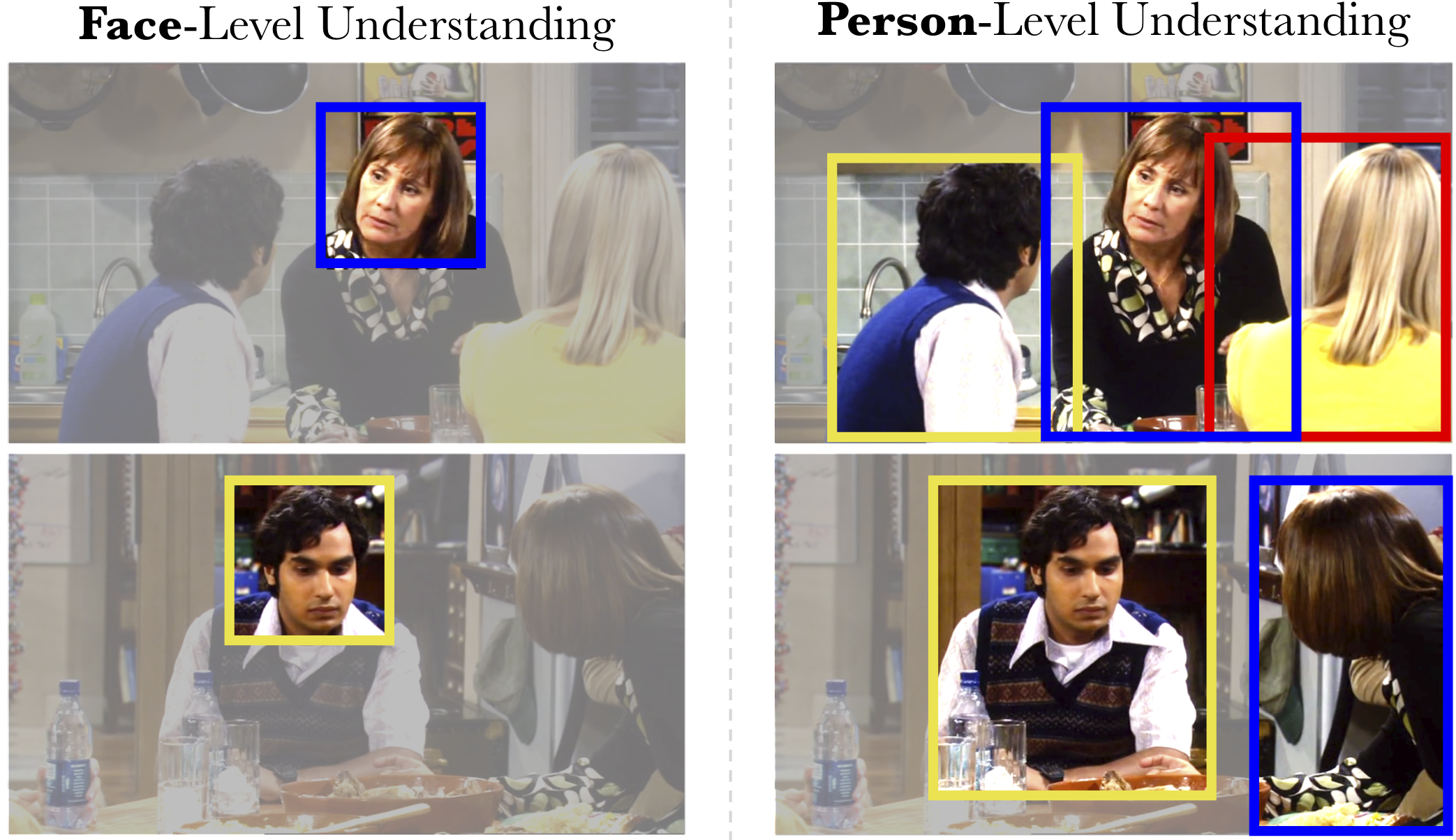} 
\end{center}
\vspace{-2mm}
  \caption{\footnotesize{
  \textbf{Video Person-Clustering -- an essential step towards story understanding.} 
Imagine trying to understand the story in the scenes above, given only the \textit{non-greyed-out} parts. Face-level understanding (left) omits important information, such as characters with their backs turned. This work addresses the new task of video person-clustering, which develops {\em person-level} understanding (right) in a scene by clustering all people, regardless of if their faces are showing or not. This is in contrast to the more limited, established,  task of face-clustering. Person-level understanding is essential for downstream applications of grouping-by-identity such as story understanding, and cannot be achieved by face-clustering alone.} 
}
\label{fig:teaser_2}
\vspace{-4mm}
\end{figure}


However, methods for clustering by identity are almost always limited to only using information from faces. Such methods have two significant drawbacks: First, they ignore many available, informative cues 
that a human would use to solve the task: (i) the person's voice
available from the audio track; (ii) the person's overall appearance (from their hair, clothes, posture); and, (iii) the editing structure (in edited material) -- such as the co-occurrence of characters in nearby shots and within a scene. 
Second, they limit the utility of clustering for downstream applications such as story understanding. 
Understanding the story-line in a scene requires knowledge of {\em all} the characters present in a scene, not just those whose faces are 
visible, \ie~ {\em person-level} not face-level reasoning.
This is illustrated in Figure~\ref{fig:teaser_2}.

Our objective in this paper is to cluster people (or more precisely person-tracks, which depict an entire body in any pose) by identity
 in movies and TV-material, as a first step towards
{\emph{story-level understanding}}. We 
cluster people, rather than just faces, and use all cues (face, voice, body appearance, editing structure), 
including tracks of people from behind without a visible face. 

To see the value and necessity of this multi-modal approach, consider the problem of determining if two poor resolution faces
depict the same person or not -- the voice can discriminatively resolve this ambiguity. Similarly, consider the problem
of determining if a person seen speaking to camera in one shot, is the same as the person seen from behind in
a following shot -- the hair and clothes can provide the link. In Figure~\ref{fig:teaser_2}, for example, how would 
the people seen from behind be identified other than by clustering their hair, clothes or voice with instances in neighboring shots?

More generally, 
modalities arising from the same person are both
{\em redundant} and {\em complementary}, 
and  can be used to address two fundamental problems in clustering:
how to obtain {\em pure} clusters (\ie containing 
tracks from a single person); and, how to
{\em merge} clusters without violating their purity (\ie by contaminating them with tracks from another person).
They can be used  to obtain very pure clusters by requiring agreement (\eg on both face and voice) in order for tracks to be grouped together; and can be used to merge clusters which could not otherwise
be confidently merged with a single modality, 
\eg by using the common voice to merge
a frontal
with a profile face cluster  (where the face descriptors of each cluster may be  different). 
In this way, multiple modalities provide a {\em bridge} between otherwise
unmergeable clusters. Methods that merge clusters using  a single modality inevitably sacrifice purity.

In this paper, we introduce a new method for the task of video
person-clustering, \methodnameMM (\mnameMM), that uses multiple
modalities -- face, voice, and body appearance. It  builds on 
recent methods that use first nearest
neighbour~\cite{finch,Kalogeiton20,JarvisP73} clustering algorithms,
and  is designed to
take advantage of the redundancy and complementarity of the
modalities, as discussed above, and to incorporate lessons from the
face-clustering literature, such as cannot-link constraints and using
the video editing structure~\cite{bauml2013semi,cinbis2011iccv} (Section~\ref{sec:method}).


To evaluate the multi-modal person-clustering task, we require a dataset with person-level annotations. 
However, there are very few such datasets due to the previous emphasis on face-clustering 
and moreover, most face-clustering and labelling datasets, such as Buffy~\cite{Everingham06a} and TBBT~\cite{TVD}, are based on TV material with limited diversity in skin color.
For these reasons, we introduce a new \datasetname (\dnameemphasis) where we: (i) 
re-purpose multiple existing face datasets by adding person-level multi-modal annotations (\eg all person-tracks and voice utterances); 
and (ii) include different TV shows and films (hereby referred to under the unified term \textit{program sets}) to address this lack of diversity. \dnameemphasis consists of visually disparate program sets,  
and includes body-tracks, face-tracks when visible; and voice utterances when speaking, for all annotated characters. We provide features so that future clustering algorithms can be compared easily and fairly (Section~\ref{sec:dataset}).


We show the effectiveness of multi-modality and outperform strong baselines for person-clustering on \dnameemphasis (Section~\ref{sub:res_personclustering}),  and explore this new expansive task for story understanding (Section~\ref{sub:res_story_understanding}). Our method also significantly outperforms the  face-clustering state of the art on both TBBT and Buffy by over 10\% NMI (Section~\ref{sub:res_faceclustering}). Note that our goal is multi-modal clustering and not representation learning. 
Thus, we do not propose a new architecture or
train a network for better features. 
Instead, we use features from pre-trained networks (for
face and speaker recognition) and only train a network where it is necessary for body Re-ID. 
A broader impact statement is included in the appendix. VPCD is publicly released and can be found here~\cite{vpcd_website}.

\section{Related Work}
\label{sec:related_work}

\noindent
In this work, we focus on multi-modal person-clustering in videos. Similar works target the more limited task of face-clustering or labelling, person Re-ID, or person search. We describe them, and also discuss similar datasets to \dnameemphasis.

\noindent 
\textbf{Face-Clustering.} A well studied task for both images~\cite{otto2017pami,ho2003cvpr,berg2004cvpr,rothlingshofer2019self} and videos~\cite{Kalogeiton20,tapaswi2014total,ball,zhang2016eccv}, with difficulty arising from the variation of pose, lighting, and emotion~\cite{lin2018cvpr,he2018aaai} in faces of the same identity. Most video approaches exploit the spatio-temporal continuity and find must-link and cannot-link clustering constraints~\cite{wagstaff2001icml,de2012constrained,bauml2013semi,cinbis2011iccv,Kalogeiton20,sharma2019self_before34,sharma2020clusteringsota,tapaswi2014total,wu2013cvpr,xiao2014eccv}, or additional constraints from the structure of videos~\cite{tapaswi2014total}. 
Most works approach face-clustering with metric or representation
learning~\cite{wu2013iccv,wu2013cvpr,cinbis2011iccv,Fitzgibbon02,sharma2019self_before34,sharma2020clusteringsota,sharma2019video34,ball}. For 
instance, \cite{ball} map features from the same identity to a
fixed-radius sphere, while Sharma \etal use supervision
from video
constraints~\cite{sharma2019self_before34,sharma2019video34} or weak
clustering labels~\cite{sharma2020clusteringsota}. These methods, however, are limited by the 
relatively small training sets available from particular TV-shows.
For this reason, some recent approaches focus on
simply clustering pre-trained features that have been learnt on very large-scale
face datasets. Sarfraz \etal~\cite{finch} propose a simple first
nearest neighbour clustering method upon pre-trained
features, FINCH, and show impressive results. More recently, \cite{Kalogeiton20} combines~\cite{finch} with spatio-temporal
constraints and improves performance. \emph{All} the above works focus
on the limited task of clustering faces (Figure~\ref{fig:teaser_2} - left),
whereas our focus is multi-modal person-clustering \ie
clustering every appearance of characters, regardless of whether their
face is visible (Figure~\ref{fig:teaser_2} - right), and using multiple modalities.

\noindent \textbf{Face-Labelling.} 
The task of classifying faces by identity - most works address this by using face-appearance with supervision from transcripts aligned to subtitles~\cite{Everingham06a,Bojanowski13,Cour10,Parkhi15a,Sivic09,Everingham09,bauml2013semi,tapaswi2012cvpr,ramanathan2014linking}, for example by using Multiple Instance Learning~\cite{haurilet2016naming,Bojanowski13,Kostinger11,Wohlhart11}. Some exploit 
cues other than faces from videos:
\cite{ramanan2007leveraging} use clothing to match faces in TV-shows across shot boundaries, while
\cite{Nagrani17b,Brown21} use face and voice to label faces.
\cite{poignant2017multimodal} use face and voice to retrieve a list of shots containing a named person, by searching for their name in subtitles and displayed text.
These works  focus only on visible faces and although some are
multi-modal (face, voice and/or text supervision), the text is typically obtained from external sources (\ie transcripts). 
Our task is different, as we cluster rather than label, and thus do not require character-classifiers or
ID supervision  or extra annotation, and we use all available cues \ie editing structure and multi-modality.

\noindent \textbf{Person Re-ID.} 
 The task of re-identifying pedestrians in CCTV  - 
 typically~\cite{zheng2015scalable,zheng2017unlabeled,li2014deepreid,wei2018person}, each body is fully visible and walking, the clothing remains constant for each identity, and the images are low resolution. This differs substantially 
from person-clustering in TV and film material, 
where there is large pose variation (\eg sitting, standing, lying down), occlusion, and the clothing frequently changes for each identity. A full literature review is out of scope. 
Closer to our task are works on person-retrieval in photo albums~\cite{zhang2015beyond, joon2015person,Sivic06a} or person-search from portraits in videos~\cite{huang2018person,xia2020online}. 
\cite{huang2018person,Sivic06a} use face and body features, while \cite{xia2020online} use audio. 
The TRECVID Instance Search challenges~\cite{Awad2019TRECVID2A}  involved retrieving a list of shots that contain an identity, given a query video for that identify.
In contrast, we cluster all characters at the track-level in videos without requiring search queries.

\noindent \textbf{Related Datasets.}
Various face-clustering datasets have been proposed~\cite{Everingham06a,TVD,Kalogeiton20,hannah,ghaleb2015accio,cour2009learning}.  These follow some similar trends:   
(a) are limited in size, consisting of a movie or some TV show episodes;
(b) under-represent most demographic groups; and
(c) contain only face annotations, so cannot be used for the broader multi-modal person-clustering task. Several story understanding~\cite{huang2020movienet,Bain20} or person-search~\cite{huang2018person} datasets with face and/or body annotations exist. These cannot be used for our task, 
as they lack audio~\cite{huang2018person,huang2020movienet} or contain only partial annotations
such as keyframes~\cite{huang2020movienet} or for a subset of tracks~\cite{Bain20}. 
Furthermore none contain labelled voice utterances.
Instead, \dnameemphasis contains 6 different TV-shows and movies, representing a diverse range of characters, and containing \textit{multi-modal annotations} for all annotated characters. 

\noindent
\textbf{Story Understanding.} 
This targets automatic understanding of human-centred story-lines in videos. It has been formulated in several ways, \eg grouping scenes by story threads~\cite{ercolessi2012stoviz,rao2020local},  learning character interactions~\cite{tapaswi2014storygraphs,Marin19a} or relationships~\cite{kukleva2020learning}, creating movie graphs~\cite{vicol2018moviegraphs}; or 
text-to-video retrieval from narrating captions~\cite{Bain20}, with several datasets~\cite{Bain20,huang2020movienet} introduced. Many  works~\cite{Bain20,vicol2018moviegraphs,kukleva2020learning} highlight the importance of knowing who is present in a scene for understanding the story. This is the focus of our work.

\ifx
\noindent \textbf{Algorithmic Bias.} 
The performance of modern face recognition and clustering systems is impressive~\cite{deng2019arcface,Kalogeiton20}, often surpassing humans~\cite{lu2014surpassing}, partly due to very large datasets~\cite{Cao18,guo2016ms,maze2018iarpa}. 
Recent studies shed to light the problem of algorithmic bias~\cite{grother2014face,feldman2015certifying,barocas2016big,leavy2018gender,datta2018discrimination,speicher2018potential,barocas2017fairness,torralba2011unbiased}. 
\cite{buolamwini2018gender} show that methods trained or evaluated with biased data result in serious implications for algorithmic discrimination, particularly for demographic groups underrepresented in the data, thus revealing the need for balanced datasets. 
In the face community, several  benchmarks~\cite{sculley2019inclusive,klare2015pushing,levi2015age} expose algorithmic discrimination, such as VGGFace2~\cite{Cao18}, a pioneer face dataset covering different types of bias including gender and geographic diversity, IJB-A~\cite{klare2015pushing}, and Adience~\cite{levi2015age}, for age and gender classification. 
However, as shown in Section~\ref{sec:dataset}, current benchmarks for person-clustering are heavily skewed towards certain demographics. 
Thus, we release a far larger, balanced benchmark, with gender labels for evaluating and exposing algorithmic bias in clustering.

\fi

\section{Method}
\label{sec:method}
Here, we describe the \methodnameMM (\mnameMM) method for person-clustering in videos. 
It is a single hierarchical agglomerative clustering~\cite{Partitional} (HAC) approach that groups person-tracks by identity using similarities of modality features, together with constraints arising from the video structure. 
\mnameMM uses pre-computed features, and hence does not require any training outside of simply learning optimal hyper-parameters, and can then run out of the box for any video dataset.
In this work, we use three modalities (face, voice, and body appearance) but \mnameMM can easily scale to any number of modalities.

\ifx
\begin{figure*}[ht!]
\begin{center}
\includegraphics[width=\linewidth]{figures/resultsv2-2.pdf}
\end{center}
\vspace{-5mm}
  \caption{\footnotesize{\textbf{High-precision cluster examples produced by \mnameMM on Sherlock from \dnameemphasis}. Stage 1 produces several high-precision clusters (grey ellipses) based only on face-appearance. This results in more clusters than characters, which consistently contain the same identity. 
  The clusters are further grouped in Stage 2 by face-body bridges (red) and face-voice bridges (green). Stage 3 clusters back of bodies (blue) in their respective clusters. The voice-bridges connect tracks that are difficult to distinguish as the same identity using face alone \eg bottom: a profile speaking Sherlock is bridged to a small frontal speaking face of Sherlock. The body-bridges are able to connect tracks where the face is in extreme profile (far left), or obscured (far right), yet the identity is wearing the exact same clothing. Typical failure cases include: low lighting, low resolution, rapid camera movements, dark identical costumes, or no voice (mixed, bottom right), where even humans find it challenging to correctly identify the characters. \todo{shorter in height and fix names}
  }
  }
  \vspace{0mm}
\label{fig:cluster_res}
\end{figure*}
\fi 

\noindent \textbf{Overview.} 
\mnameMM consists of three stages. 
\textbf{Stage~1} creates high-precision clusters using a single modality, here the face. 
We group person-tracks that share a first nearest neighbour (NN) using multiple iterations of HAC 
(as in~\cite{finch,Kalogeiton20,JarvisP73}).
We follow this trend  
 subject to two additional \emph{constraints}: a cannot-link constraint for concurrent tracks (as in~\cite{Kalogeiton20} 
based on~\cite{bauml2013semi,cinbis2011iccv}), and a 
conservative threshold on the
maximum NN distance.  This results in $K_1$ clusters
(Section~\ref{sub:stage1}).
\textbf{Stage~2} exploits 
multi-modality to \emph{bridge} clusters that were otherwise unmergeable by the single face  modality with a conservative threshold; 
in particular, by requiring that different modalities (\ie face and voice) concur on the merge (Section~\ref{sub:stage2}). 
\textbf{Stage~3} clusters tracks without visible faces, and hence that are not yet clustered by the first two stages.
Constraints from the editing structure (neighboring shots)
and a conservative threshold on body features (so that they depict
the same person with the same clothing) are used to link face-less
person-tracks to clusters with faces (Section~\ref{sub:stage3}). 
%
Here, we describe the stages, algorithm design choices, and how the hyper-parameters are learnt. The method is visualised in Figure~\ref{fig:cluster_res}.

\noindent \textbf{Notation.}
Given a dataset with person-tracks and $C$ characters, where $x_{i}$ is a single person-track, 
the goal is to cluster all $x_{i}$ by identity into $C$ clusters ($C$ is unknown). 
Each 
person-track $x_{i}$ 
is represented by one feature vector per available modality, 
\ie $x{=} \{x_{f}, x_{v}, x_{b}\}$, with $x_{f}$, $x_{v}$, $x_{b}$ the face, voice and body-track features, respectively. 
The availability of each 
feature vector is dependant upon the part of the person that is visible (face and/or body), and if they are speaking. 
For each person, at least one of $x_{f}$, $x_{b}$ are available.  
Let  $d(x_{i},x_{j})$ be the distance between two track features of the same modality, and $d_f, d_v$ and $d_b$ the distances between two face, voice or body-tracks, respectively; the lower the value, the more likely the tracks depict the same identity. 
NN is nearest neighbor; $n_{x_i}^1$ is the first NN track of track $x_{i}$. The set of video frames that $x_{i}$ is present in is denoted by $T_i$.

\subsection{Stage 1: High-Precision Clustering}
\label{sub:stage1}

Stage 1 creates  high-precision clusters, each containing tracks of the same identity. It uses only the face modality
as this is the most discriminant of the three (face, voice and body), and thus is least likely to group different identities in the same cluster. Here, we use a NN clustering method~\cite{finch,Kalogeiton20}, subject to two clustering constraints.

\noindent
\textbf{Clustering Constraints.} 
A NN is only considered valid if the resulting merge satisfies:
\textit{(1) A Spatio-Temporal Cannot-link Constraint:} Tracks that have (partial) temporal overlap cannot be grouped together, since they must represent different characters as they appear together 
in at least in one frame (introduced by~\cite{Kalogeiton20});
and \textit{(2) A NN Distance Constraint:} the
distance $d_f(x_{i},n_{x_i}^1)$ between a track $x_{i}$ and its first NN $n_{x_i}^1$ 
is less than a strict threshold $\tau_f^{\text{tight}}$ for Stage 1.

\noindent
\textbf{Clustering process.}
At every iteration (cluster partition $\Gamma$), each cluster is
grouped with its NN cluster, \ie the closest. Specifically, the first
partition groups tracks into clusters through first NN relations,
while following partitions group the clusters formed in the
previous partition; each cluster is represented by the average
of the features it contains.  Following the notation
of~\cite{finch}, at each partition $\Gamma$, the method forms $K_\Gamma$
clusters by merging tracks that are either first NN (mutually or one
is the first NN of the other) or have a common NN $n_{x_i}^1$, as
described by the adjacency matrix: 

\vspace{-5mm}
\begin{equation}
\begin{footnotesize}
A(x_i,x_j) =
\begin{dcases}
1 & \begin{array}{@{}l} 
\displaystyle
\phantom{{}-}
\text{if } \left( x_j = n_{x_i}^1 \text{ or }  n_{x_j}^1 = x_i \text{ or } n_{x_i}^1 =n_{x_j}^1  \right) \\
\qquad{}
\text{ and } T_i \cap T_j = \emptyset \text{, } d_f(x_{i},n_{x_i}^1) \leq \tau_f^{\text{tight}}\\

\end{array} \\
0& \text{otherwise} \text{.}
\end{dcases} 
\end{footnotesize}
 \label{eq:finch}
\end{equation}

\vspace{-1.5mm}
\noindent
\textbf{Discussion.}
In standard HAC the clustering continues until all clusters merge to one. 
Including the constraints  introduces strict stopping criteria, and therefore the clustering stops when either the clusters are all more than a distance $\tau_f^{\text{tight}}$ apart, or they are separated by a cannot-link constraint. This results in ${K_{1}}$  high-precision clusters, where we expect ${K_{1}} \geq C$.
The very simple addition of a distance threshold 
leads to a significant improvement in clustering results over~\cite{Kalogeiton20,finch,ball} 
(Section~\ref{sub:res_faceclustering}). Without this constraint, little
prevents an incorrect merging of clusters of different identities and
the subsequent creation of low-precision clusters. 

\subsection{Stage 2: Multi-modal Cluster Bridging}
\label{sub:stage2}

Combining a discriminative modality with the constraints results in \textit{high-precision} clusters. 
However, a single modality alone cannot continue making confident merges without sacrificing purity. 
Thus, Stage 2 merges these clusters by exploiting multiple modalities \ie face and voice. 

\noindent \textbf{Modality-pair merges.} \new{To further merge clusters, we demand that two modalities agree that the clusters contain the same identity.}
Therefore, we require that the distances for the face and voice are both below new thresholds, \ie $d_f{<}\tau_f^{\text{loose}}$ and $d_v{<}\tau_v^{\text{loose}}$. Note, here we use features taken from tracks within clusters, rather than averaged cluster features.  
$\tau_f^{\text{tight}}$ is raised by just a small margin, $\delta$, \ie $\tau_f^{\text{loose}}{=}\tau_f^{\text{tight}} + \delta$, due to the concurrent agreement from the voice.
 
\noindent \textbf{Discussion.}
This stage results in $K_{2}$ clusters with high-precision, where $K_{2} \leq K_{1}$.  
Here, we use face and voice as they have been shown to be coupled~\cite{Nagrani18a,Nagrani18c} and to contain redundant, identity discriminating information. \new{An alternative is to require that the voice modality alone provides a confident (\ie tight threshold) match, \eg  two person-tracks with the same voice. We find however that voice alone cannot reliably join clusters of the same identity. This can be because two identities with the same emotion in their voice (\eg shouting, crying) can appear similar to the less discriminative voice embedding (more in the appendix). } 
\vspace{-3mm}




\begin{figure}[t!]
\begin{center}
\includegraphics[width=\linewidth]{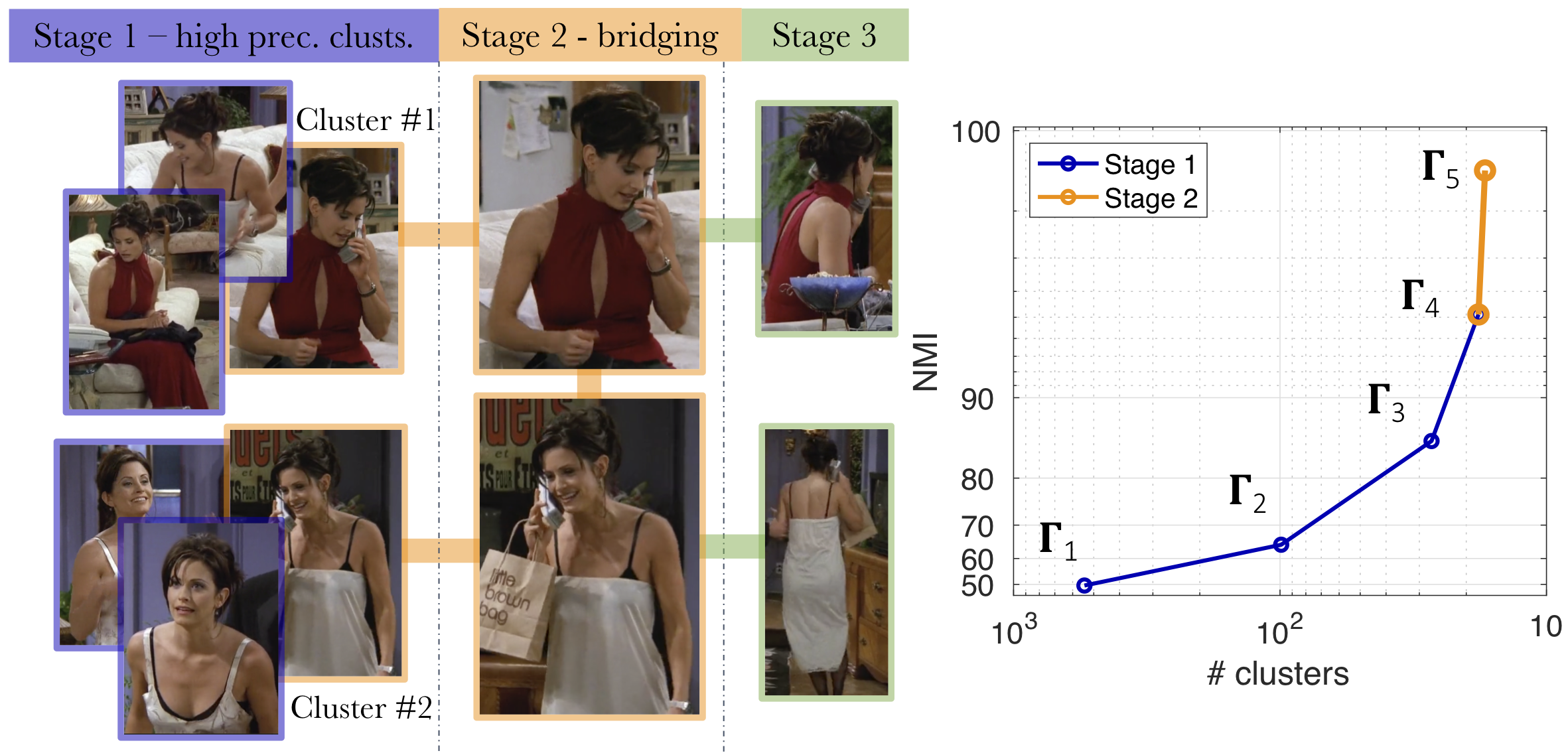}
\end{center}
\vspace{-5mm}
  \caption{\footnotesize{\textbf{The clustering process of \mnameMM.} (Left) Example person-tracks at each stage of \mnameMM. Two high-precision clusters from Stage~1 depicting the same character. One contains near-frontal faces (below) and one profiles (top), hence the single face modality cannot confidently merge the two. Stage 2 uses a talking person-track from each cluster to form a bridge, by demanding the agreement of both face and voice modalities that these contain the same identity. Stage 3 merges face-less bodies into the formed cluster. (right) The NMI and number of clusters at each partition, $\Gamma$, of stages 1 and 2 on an example video from \dnameemphasis. At each partition the number of clusters decreases, while the normalised mutual information increases. At $\Gamma_4$ Stage 1 clustering stops. Stage 2 progresses to $\Gamma_5$ by bridging clusters. Stage 3 does not affect the number of clusters.}
  }
  \vspace{-2mm}
\label{fig:cluster_res}
\end{figure}

\subsection{Stage 3: Clustering backs} 
\label{sub:stage3}

Stages 1 and 2 result in high-precision person clusters.
Nevertheless, they do not account for person-tracks with no visible
face, for instance when viewed from behind, \ie a \emph{face-less}
person. The goal of Stage 3 is to add the face-less person-tracks into
their respective high-precision clusters using the modality of
body-appearance. Here, we use the editing structure of the videos,
given that the appearance of the same character can change
dramatically between scenes. As discussed above, body features may not
be discriminative for identity if characters
are wearing very similar clothing. We determine such body-tracks
using the simple ratio-test introduced
in~\cite{lowe2004distinctive}. Specifically, for each body-track we
compute the first and second NN distances, $d_{b,x_i}^1$ and
$d_{b,x_i}^2$. If the ratio, $d_{b,x_i}^2 / d_{b,x_i}^1$ is higher than
a threshold $\rho$ then the body-track is classified as non-distinctive and is ignored.

In detail, for assigning face-less people to clusters, we find the NN body-track (that has a face and therefore is already clustered) that does not violate the ratio-test in a neighbouring shot, and assign the face-less person to this cluster. Given that the same person is most likely wearing the same outfit in the same or neighbouring shots, we only examine the distance between body-tracks from these shots. At this stage, some backs cannot be clustered with high confidence, either because they are not similar to any nearby body or because they themselves fail the ratio test for being a non-distinctive feature. Our design choice is to ignore these backs. In detail, we ignore any back for which the NN distance is more than a threshold  $\tau_b^{\text{back}}$. Note, this stage does not change the number of clusters from $K_{2}$.

%

\noindent \textbf{Required Number of Clusters.}
Suppose we know the number of characters $C$, and hence the number of clusters. 
Our goal is to reduce $K_{2}$ to the desired $C$ (typically $K_{2} \geq C$).
Previous methods~\cite{ball} employ HAC; however, this suffers from reliance on features that can no longer confidently discriminate between clusters of the same person.
Instead, we employ a cluster prior: 
there is no identity overlap amongst the largest clusters \ie they contain unique identities, and conversely there is likely an identity overlap between a small and large cluster. 
Our intuition is that big clusters contain ample information about an identity, and consequently if two large clusters contained the same identity, then they would have been merged. 
Therefore, we iteratively merge the smallest with the largest cluster until there are $C$ clusters. In practice, we observe that small clusters 
contain blurry or low-resolution tracks, and so could not confidently be merged at earlier stages.

\noindent \textbf{Discussion.}
Most methods~\cite{ball,sharma2019self_before34,sharma2020clusteringsota} fine-tune character features on a video dataset. 
Instead, \mnameMM operates on pre-trained features, thus reducing the computational burden and leading to increased generalisation capabilities. 
An extension would be to replace the constraints with a cost function optimisation approach, allowing a cannot-link to be correctly broken for a person's reflection in a mirror. 


\subsection{Learning Hyper-Parameters}
\label{sub:hyper-param}
The hyper-parameters for \mnameMM are learnt on the validation partition of \dnameemphasis. The visually disparate program sets in the test partition are disjoint from those in the validation, yet these parameters are kept constant. For the hyper-parameter associated with the face modality ($\tau_f^{\text{loose}}$) this is possible as the face features are trained on millions of faces~\cite{Cao18}, and therefore are highly discriminative and
universal (generalise well across different program sets). However, voice identity features are less universal than face features, and hence there is not a single good choice for $\tau_v^{\text{loose}}$ that would generalise across the audibly disparate program sets. 
Instead, we learn a unique value \emph{automatically} for each. 
Our goal is to choose $\tau_v^{\text{loose}}$ to be lower than the minimum distance between voices from different people. The cannot-link constraints automatically provide face-track pairs of different identities. 
We measure the distances between different people's voices.
In practice, there are too few constraints between speaking faces to provide an accurate representation of the negative distances, as rarely two face-tracks speak in the same shot. We combine the cannot-link speaking face-tracks with clusters from Stage 1 to provide more examples. This leads to many negative distances and an accurate representation of their distribution. We select $\tau_v^{\text{loose}}$ as the lower 99.9 percentile of these distances. This provides a robust automatic threshold measure. For program sets 
with similar sounding characters, this process gives a low $\tau_v^{\text{loose}}$ (\eg Buffy -- many similar sounding teenagers).

\section{Video Person-Clustering Dataset}
\label{sec:dataset}
\begin{figure*}[ht!]
\begin{center}
\includegraphics[width=\linewidth]{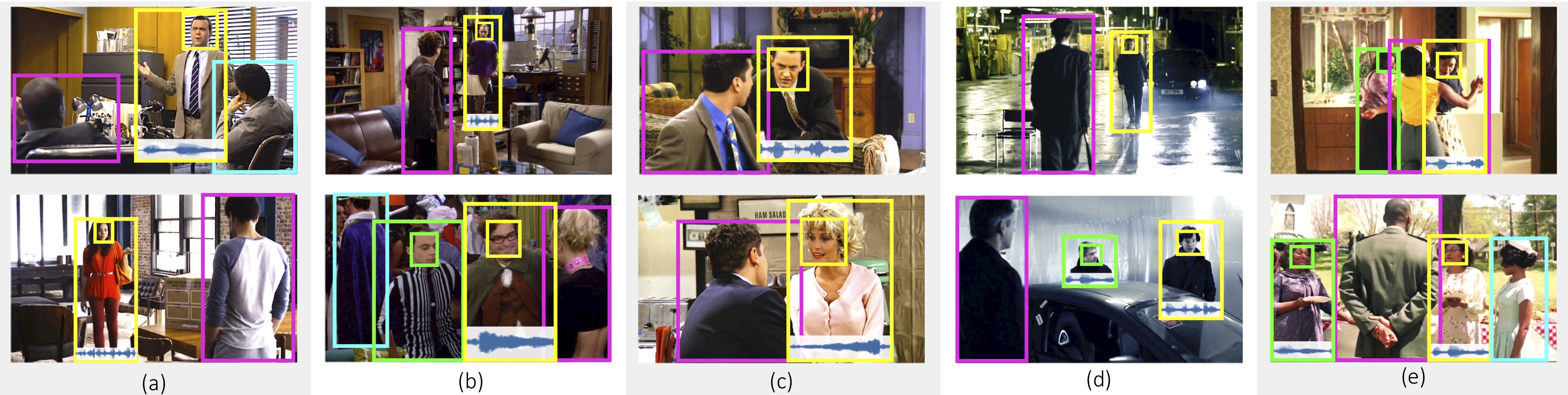}
\end{center}
\vspace{-4mm}
  \caption{\footnotesize{\dnameemphasis dataset. 
  It consists of different and diverse TV shows and movies; here, we display a subset of them: (a) About Last Night, (b) TBBT, (c) Friends, (d) Sherlock, (e) Hidden Figures.  
  \dnameemphasis contains face, body and voice tracks annotated for many characters. 
  Here, we display such examples. Each face-body pair is displayed with a unique color.  
   A diverse and representative range of characters are captured in a variety of scenes (\eg dark (d)), viewpoints (\eg (e)); and poses, including backs of bodies (magenta, cyan). When speaking, we also include a voice-track (blue signal below body-tracks).
  }}
\label{fig:dataset}
\end{figure*} 

\begin{table}[t]
\centering
\resizebox{\linewidth}{!}{
\footnotesize
\begin{tabular}{lrlrcrrr}
\toprule
\multirow{2}{*}{\textbf{Dataset}}          & \multirow{2}{*}{\textbf{\#eps}} & \multirow{2}{*}{\textbf{length}} & \multirow{2}{*}{\textbf{\#IDs}} & \textbf{Gender} & \multicolumn{3}{c}{\textbf{\#Tracks}} \\
& & &  & F/M &body & face & voice \\  
\midrule 
TBBT~\cite{TVD}     & 6  & 2h 6m    &  103   &  53/50 &  4,276   &   3,908 & 1,047\\
Buffy~\cite{Everingham06a}      &  6 &        4h 9m    &      109              &   37/70 &   7,561 & 5,832 & 1,835    \\
Sherlock~\cite{Nagrani17b}                       &       3               &      4h 30m           &           31          &     16/15  &    6,232           &       6,247 & 1,615              \\
Friends~\cite{Kalogeiton20}                       &        25              &     9h 22m            &         49            &      23/26  &   18,360             &           17,333 & 3,961               \\

ALN~\cite{somandepalli2020multi}               &        1             &       1h 40m          &          10            &    4/6  &  1,932            &         1,614 & 404           \\
HF~\cite{somandepalli2020multi}                 &         1             &     2h 7m            &           24        &  11/13  &        1,416            &     1,463 & 303    \\ \midrule 
\textbf{\dnameemphasis} &                      &      23h 54m           &        326             &                      &           39,777  & 35,396 & 9,165  \\ \bottomrule
\end{tabular} 
}
\vspace{-2mm}
\caption{\footnotesize{\textbf{\datasetname statistics.} For each program set in \dnameemphasis we detail video and annotation statistics. \#eps: number of episodes; \#IDs: number of unique characters; TBBT: The Big Bang Theory; (movies) ALN: About Last Night; HF: Hidden Figures. We cite the first published work that used each respective program set for face-clustering, but we provide additional full multi-modal annotations for  each. }} %
\label{tab:meta_dataset_stats}
\vspace{-3mm}
\end{table}

\noindent
In this section, we describe the dataset
(Section~\ref{sub:dataset_content}), the annotation (Section~\ref{sub:annotation_process}), and the feature
extraction processes (Section~\ref{sub:feature_extraction}).
%
The dataset is built on top of existing video datasets that have face-level annotations (labeled face-tracks) by adding
and annotating body-tracks, and annotating voice utterances. This is for three reasons: first, it enriches the
existing dataset by raising them to have person-level annotations; second, it enables comparisons on face-level clustering
with prior work on these datasets; and third, it means that the video material is already publicly available and we need only
release the new annotations (and features).

\subsection{\dnameemphasis content}
\label{sub:dataset_content}
\dnameemphasis contains  \textit{full multi-modal annotations} 
for primary and
secondary characters
for a range of diverse and visually disparate TV-shows and
movies 
(statistics in Table~\ref{tab:meta_dataset_stats}, examples in  Figure~\ref{fig:dataset}). 
\dnameemphasis contains annotations for 39,777
body-tracks, 35,396 face-tracks for whenever the face is visible, and 9,165
manually annotated voice-tracks for whenever each of them are
speaking. 
Identity
discriminating features (embeddings from deep networks) are provided for 
all modalities. 
A total of 23 hours of video  
cover a range
of genres and styles such as Hollywood Drama (Hidden
Figures, 2016), Romance (About Last Night, 2014), fast-paced
Action/Mystery (Sherlock, Buffy) and live studio-audience sitcoms
(Friends, TBBT). A large variety of characters are annotated,
ranging from small casts shown over many episodes (\eg Friends) to
program sets with a long-tailed distribution with many
secondary/background characters (\eg Buffy). \dnameemphasis is by far the largest dataset of its kind. The program sets were chosen such that
\dnameemphasis is representative of the diversity of people's
appearance in  the real world. There is a val. set and a test set - these are disjoint. The val. set is the first five episodes of Friends.

\subsection{Annotation Process}
\label{sub:annotation_process}
Here, we describe the annotation process for the face, body, and voice
tracks in \dnameemphasis. 
For all component program sets, the face annotations already exist, and define the characters of interest for that video. 
Our goal is to annotate their body and voice-tracks. Very often in videos, a character is seen facing from behind (Figure~\ref{fig:dataset}). This means that the existing face-tracks cannot be used to trivially annotate the body-tracks by spatial overlap (since there will be no face-track). 
We therefore combine automatic and manual annotation methods 
(more details in the appendix).


\noindent \textbf{Face.} 
We use the same face bounding-box/track annotations and ID labels as were provided with the original datasets so that we can compare to previous works on face-clustering.


\noindent \textbf{Body.} 
We detect bodies with a Cascade R-CNN~\cite{cai2018cascade} trained on MovieNet~\cite{huang2020movienet} and form tracks with an IOU tracker. When a body-track clearly corresponds to a face-track (\ie no significant IOU with any other face-track), the body-track is automatically annotated with the character name of that face-track. 
We manually annotate the remainder as well as the body-tracks corresponding to characters from behind.

\noindent \textbf{Voice.} 
We manually segment the audio-track into the speaking parts for all annotated characters. To ensure the correctness of the segmentation, the audio track was first segmented by one human annotator, and then verified by different ones. 


\begin{table*}[ht!]
\centering

\bgroup
\def\arraystretch{1.1} 
{\setlength\tabcolsep{1.1pt} 
\resizebox{\linewidth}{!}{
\footnotesize{
\begin{tabular}{lccc|cccc |cccc |cccc |cccc |cccc |cccc |cccc}
\toprule
\multirow{2}{*}{\#} &  \multicolumn{3}{c|}{\multirow{2}{*}{Modality}}   &
\multicolumn{3}{c}{\cellcolor{lavenderblue}{}  }       &\cellcolor{lavenderblue}{\textbf{$\#C_s${=}}}   
& \multicolumn{3}{c}{\cellcolor{lavenderblue}{ }}        &          \cellcolor{lavenderblue}{\textbf{$\#C_s${=}}}                                                 &
\multicolumn{3}{c}{\multirow{2}{*}{\cellcolor{lavenderblue}{}}}  &  \cellcolor{lavenderblue}{\textbf{$\#C_s${=}}} & 
\multicolumn{3}{c}{\multirow{2}{*}{\cellcolor{lavenderblue}{}}} & \cellcolor{lavenderblue}{\textbf{$\#C_s${=}}} & 
\multicolumn{3}{c}{\multirow{2}{*}{\cellcolor{lavenderblue}{}}} & \cellcolor{lavenderblue}{\textbf{$\#C_s${=}}} & 
\multicolumn{3}{c}{\multirow{2}{*}{\cellcolor{lavenderblue}{}}} & \cellcolor{lavenderblue}{\textbf{$\#C_s${=}}} & 
\multicolumn{3}{c}{\multirow{2}{*}{\cellcolor{lavenderblue}{}}} & \cellcolor{lavenderblue}{\textbf{$\#C_s${=}}} \\
& & &  &\multicolumn{3}{c}{\multirow{-2}{*}{\cellcolor{lavenderblue}{\textbf{TBBT}}}}  & \cellcolor{lavenderblue}{130}
& \multicolumn{3}{c}{\multirow{-2}{*}{\cellcolor{lavenderblue}{\textbf{Buffy}} }}        &         \cellcolor{lavenderblue}{165}                                              &
\multicolumn{3}{c}{\multirow{-2}{*}{\cellcolor{lavenderblue}{\textbf{Sherlock}}}}  &  \cellcolor{lavenderblue}{50} & 
\multicolumn{3}{c}{\multirow{-2}{*}{\cellcolor{lavenderblue}{\textbf{Friends}}}} & \cellcolor{lavenderblue}{239} 
&  \multicolumn{3}{c}{\multirow{-2}{*}{\cellcolor{lavenderblue}{\textbf{Hidden Figures}}}}   & \cellcolor{lavenderblue}{10}
&  \multicolumn{3}{c}{\multirow{-2}{*}{\cellcolor{lavenderblue}{\textbf{About Last Night}}}}   & \cellcolor{lavenderblue}{24}
&  \multicolumn{3}{c}{\multirow{-2}{*}{\cellcolor{lavenderblue}{\textbf{Average}}}}   & \cellcolor{lavenderblue}{618} 
\\ \midrule

&  F & B & V    & 
WCP   & NMI                  & CP                 & \multicolumn{1}{c|}{CR} &  
WCP                  & NMI                  & CP                 & \multicolumn{1}{c|}{CR} &  
WCP                  & NMI                  & CP                 & \multicolumn{1}{c|}{CR} & 
WCP                  & NMI                  & CP                 & \multicolumn{1}{c|}{CR} & 
WCP                  & NMI                  & CP                 & \multicolumn{1}{c|}{CR} & 
WCP                  & NMI                  & CP                 & \multicolumn{1}{c|}{CR} & 
WCP                  & NMI                  & CP                 & CR \\ 
\hline 
B-ReID  &  & {\checkmark} &  &  80.5            &  69.7                &    49.6      & 55.0 & 65.0                  &      60.9          &     52.7             & 46.8 &  61.2               &         28.9    &  43.6             & 44.3  &    70.9         & 60.4                  & 71.0                  & 56.3  &     32.6      &     23.4           &        36.8           & \multicolumn{1}{c|}{19.6}    &      41.0           &   14.1               &    37.4           & \multicolumn{1}{c|}{32.6}    & 58.5       & 42.9  & 48.5 & 42.4   \\
B-C1C &  {\checkmark} & {\checkmark} & &       87.7          & 69.2            &    39.4           & \multicolumn{1}{c|}{50.6}    &     73.6            &   58.2                          & \multicolumn{1}{c}{34.6}    &   41.6          &       77.7                            & \multicolumn{1}{c}{41.6}    &       29.3            &      43.6         &     85.3                  & 77.1  & 69.5 & 70.8 & 
        76.2      &         69.8       &        55.2          & \multicolumn{1}{c|}{50.3}    &          94.4           &      85.8      & \multicolumn{1}{c}{68.0}    &                   76.8    &         82.5          &  67.0 &  49.3 & 55.6
\\ \midrule
\mnameMM--  &   {\checkmark} &  &  &   93.5    &     84.6            &   76.4            & \multicolumn{1}{c|}{77.6}  &      80.0          &     66.7         &  63.8          & \multicolumn{1}{c|}{65.2}  &        83.8        &   52.3        &   51.2          & \textbf{58.4} &  85.7                 &      73.7            &   81.3             & 79.0 &   \textbf{77.6}   &  \textbf{70.4}          & \textbf{59.1}               & \textbf{52.1} &     95.7            &     89.7            &    98.2       &      \multicolumn{1}{c|}{86.3}    &  86.1     & 72.9 &  71.7 &    69.8   \\
$\mnameMM_v$ & {\checkmark} &    &  {\checkmark} &       93.5       &     84.6             &  76.4                   & \multicolumn{1}{c|}{77.6}    &  80.1  &    67.2    &  64.2         &  64.7                &    84.5  & 59.3               &    54.9           & \multicolumn{1}{c|}{57.3}     &        86.9           &    75.3         &  84.0        &  82.8  &  77.6
             & 70.4            &   59.1   & \multicolumn{1}{c|}{52.1}   &          96.0        &\textbf{90.5}          &         98.3        & \multicolumn{1}{c|}{86.4}  &    86.4   & 73.5 & 72.3 &    69.7          
\\
$\mnameMM_b$  & {\checkmark} &{\checkmark}  &  &  \textbf{96.9} &   \textbf{92.8}      &\textbf{80.4}  & \textbf{79.6}  &    85.7   &     75.6          &         68.1      & \textbf{67.9}    &    84.1       &   52.9               & 51.7            &  54.3  &   89.5          &   81.3          &  84.6           &  82.4 & 
77.6          &         70.3        &   59.0               & 52.0 &   95.7 &   89.4       &    98.2          &     86.3        &  88.2    & 77.1        & 74.0 & 70.0 
\\  \hline 
\mnameMM &  {\checkmark} &{\checkmark}   &{\checkmark}   &   96.9          &    92.8  &     80.4      &   79.6  &     \textbf{85.8}          &   \textbf{76.4}      &  \textbf{68.4}               &  67.2 &    \textbf{84.8}      &        \textbf{60.0}         &       \textbf{55.2}       & 57.2 &             \textbf{90.8}  & \textbf{83.1}                &           \textbf{87.7}  & \textbf{86.6}  &  77.6
           &        70.3      &     59.0     & 52.0 &  \textbf{96.0}      &        90.2          &       \textbf{98.3}            & \textbf{86.4} &    \cellcolor{mistyrose}{\textbf{88.6}}          & \cellcolor{mistyrose}{\textbf{78.8}} & \cellcolor{mistyrose}{\textbf{74.8}} & \cellcolor{mistyrose}{\textbf{71.5}}           
 \\
\bottomrule
\end{tabular}
}
}
}
\vspace{-1mm}
\caption{\footnotesize{\textbf{Person-Clustering Results on \dnameemphasis.} For each program set, each metric is averaged across all episodes. AT protocol. The `Average' column reports averaged metrics across all six program sets. $\#C_s$ is the sum of ground truth clusters across each episode in each program set. We report two strong baselines (B-ReID, B-C1C, Section~\ref{sub:res_personclustering}) and an ablation on the modalities used. Keys: F-face, B-body, V-voice. \textit{Modality}: used modalities.} }
\label{tab:person_results}
\egroup
\end{table*}

\subsection{Feature Extraction}
\label{sub:feature_extraction}
\noindent \textbf{Face.}  
We use L2-normalised 256D features, extracted from an SENet-50~\cite{hu2018squeeze} pre-trained on MS-Celeb-1M~\cite{guo2016ms}, and fine-tuned on VGGFace2~\cite{Cao18} (same as~\cite{TVD,Everingham06a,Nagrani17b,Kalogeiton20}).


\noindent \textbf{Body.}
For all body detections, we extract 256D features with ResNet50~\cite{He16} trained on CSM~\cite{huang2018person}. 
We average the features across each body-track, and then L2-normalise them.

\noindent \textbf{Voice.} 
Following~\cite{chung2020in}, we extract a single, L2-normalised 512D speaker embedding from each voice segment using a thin-ResNet-34~\cite{Xie19a,He16} trained on VoxCeleb2~\cite{Chung18a}.

\begin{figure*}[htb]
    \centering 
\begin{subfigure}{0.8\textwidth}
  \includegraphics[width=\linewidth]{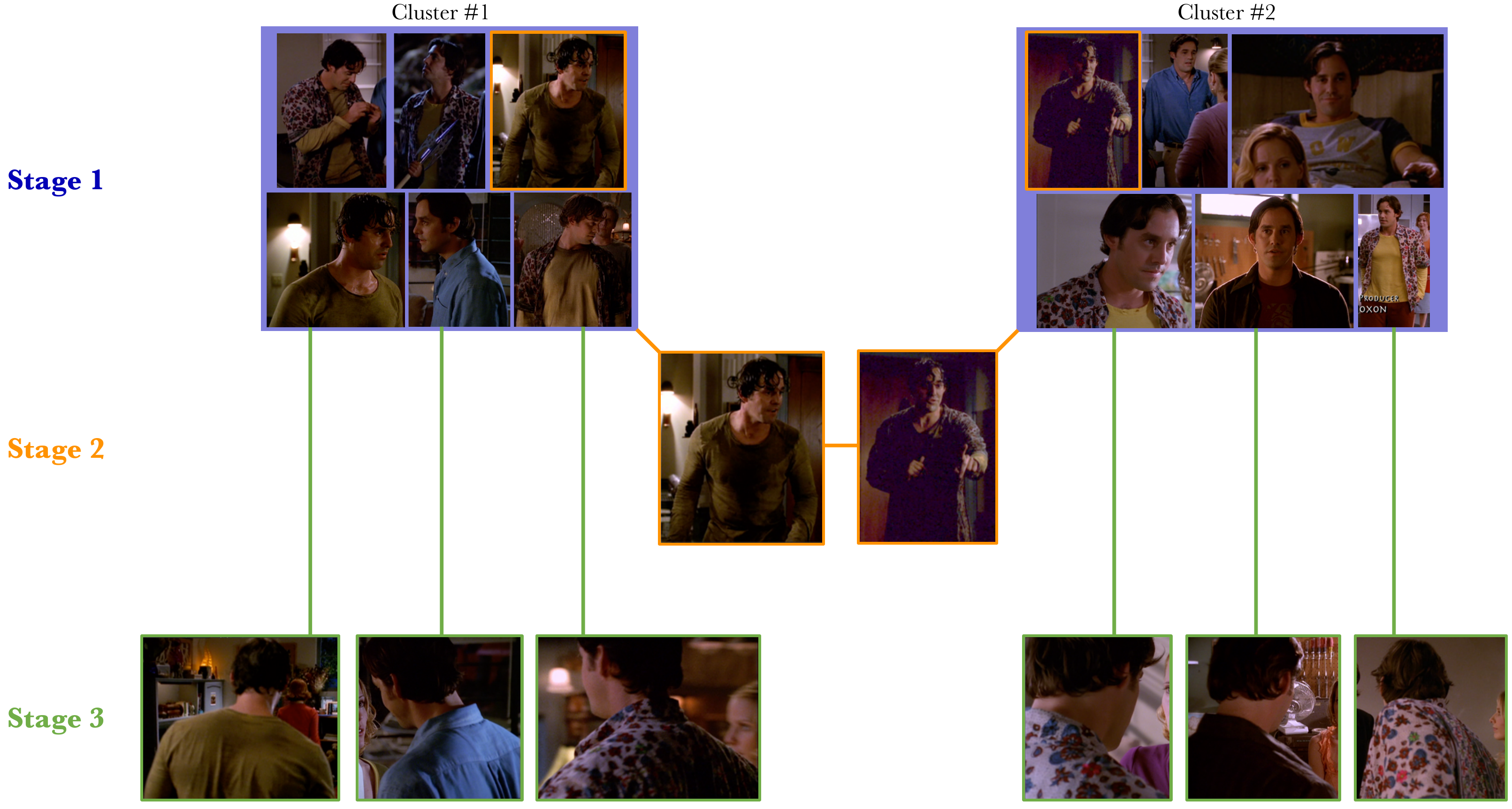}
  \vspace{1mm}
  \caption{\footnotesize{\textbf{Clustering Process of \mnameMM for a character in Buffy.} Stage 1 produces high-precision clusters. Cluster \#1 contains mainly profile and downwards-facing views of the character, while Cluster \#2 contains frontal facing views. Both clusters contain very different clothing and body poses. The face modality alone can no longer confidently merge these clusters. Stage 2 merges the two clusters using multi-modal bridges between a speaking person-track from each cluster. Stage 3 then merges back views into these clusters via body features. Back views of the character are merged via frontal appearances in nearby shots where the character is wearing the same clothing.}}
  \label{fig:clustering1}
  \vspace{1mm}
\end{subfigure}\hfil 
\begin{subfigure}{0.8\textwidth}
  \includegraphics[width=\linewidth]{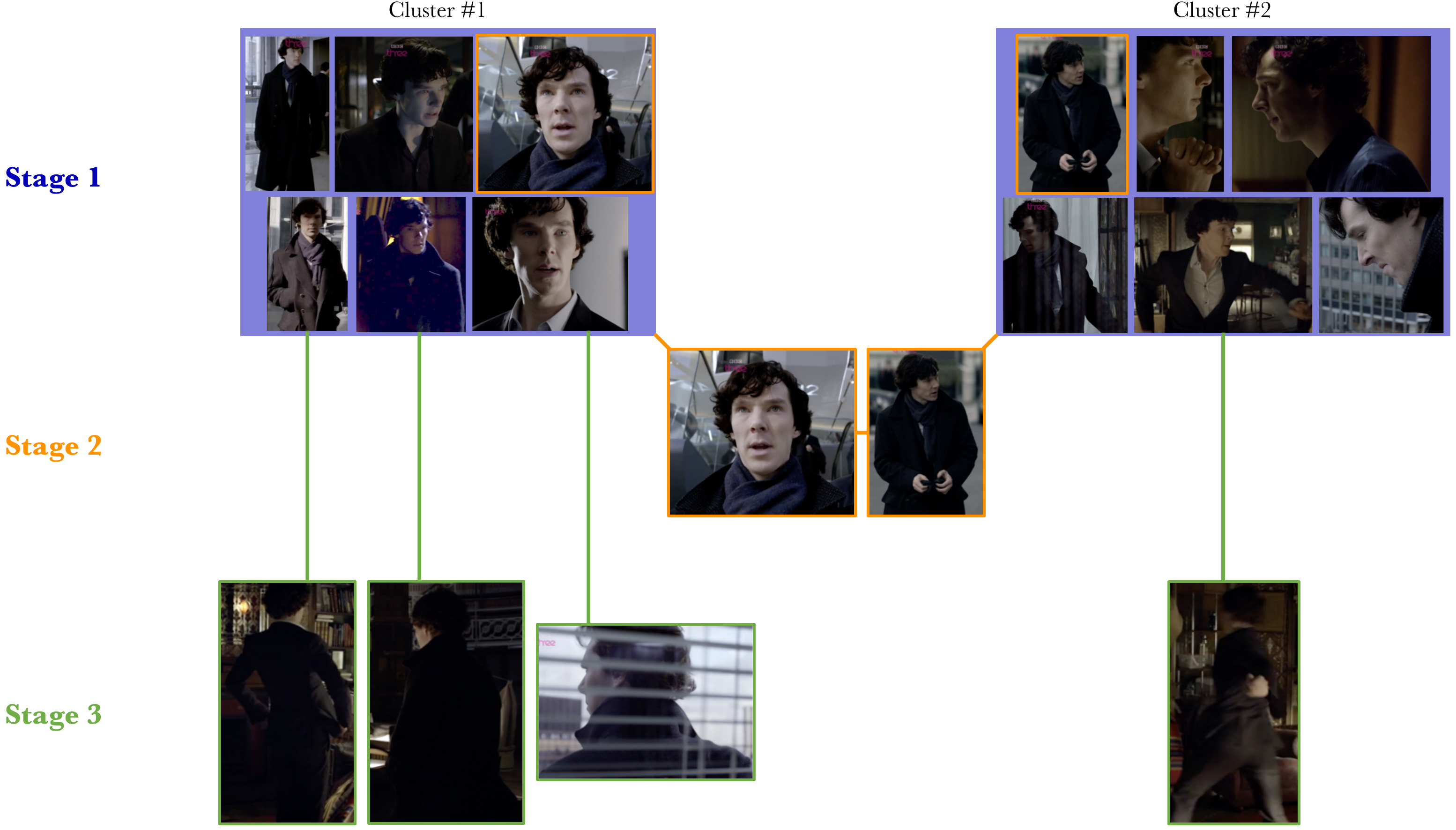}
  \vspace{1mm}
  \caption{\footnotesize{\textbf{Clustering Process of \mnameMM for a character in Sherlock.} Stage 1 produces high-precision clusters. Cluster \#1 contains mainly frontal face views, while Cluster \#2 contains profile face views. Both clusters contain very different lighting conditions, body poses; and  camera-views of the same character. Stage 2 merges the two clusters where the face alone could not, by using multi-modal bridges between a speaking person-track from each cluster. Stage 3 then merges back views into these clusters via body features. Back views of the character (both full-body, and over-the-shoulder views) are merged via frontal appearances in nearby shots where the character is wearing the same clothing.}}
  \vspace{1mm}
  \label{fig:clustering2}
\end{subfigure}\hfil 
\caption{\footnotesize{\textbf{Clustering Process of \mnameMM.} For two program sets from \dnameemphasis,  (a)-Buffy,  and (b)-Sherlock, we show the clustering process for one of the principal characters.}
}
\label{fig:qual_clust}
\end{figure*}

\section{Experiments}
\label{sec:results}
Here, we evaluate \mnameMM. We first give experimental details, followed by person-clustering results on \dnameemphasis and provide ablations. 
We compare to previous face-clustering works and finally examine the advantages of person-clustering for story understanding.
Further ablations and experiments on clustering all characters in all videos 
simultaneously are included in the appendix. 

\noindent
\textbf{Implementation details.} We use the face, body and voice track annotations and features from \dnameemphasis 
(Sections~\ref{sub:dataset_content},\ref{sub:feature_extraction}). 
For all modalities, feature distances $d_f$,$d_b$,$d_v$ are computed using (1 - cosine similarity). 
As described in Section~\ref{sub:hyper-param}, parameters are learnt on the \dnameemphasis val.\ set. The values are:
$\tau_f^{\text{tight}}{=}{0.48}$,  $\delta{=}0.025$, $\rho{=}0.9$ and $\tau_b^{\text{back}}{=}0.4$. 
These parameters are fixed for all experiments, and only have to be re-learnt  
if the features  change. Details on the automatically selected $\tau_v^{\text{tight}}$ values are in the appendix.

\noindent \textbf{Metrics.}
For each dataset in \dnameemphasis, we measure each metric at the episode level and  average over all episodes. 
Following~\cite{Kalogeiton20,ball}, we use Weighted Cluster Purity (\textbf{WCP}) and Normalized Mutual Information (\textbf{NMI}). 
\noindent WCP weights the purity of a cluster by the number of tracks belonging in it. 
NMI~\cite{manning2008introduction}  measures the trade-off between clustering quality and number of resulting clusters.
\noindent \textbf{Character Precision and Recall (CP, CR)} are computed using the number of ground truth identities. Each identity is uniquely assigned to a cluster. CP is the proportion of tracks in a cluster that belong to its assigned character, while CR is the proportion of that character's total tracks that appear in the cluster. They are averaged across all characters, thus weighting each equally.

\noindent
\textbf{Test protocol.}
We evaluate:  (i) automatic termination (AT), \ie unknown number of clusters, and (ii) oracle cluster (OC), when known. 
AT is realistic for applications, while OC offers a fair comparison to the state of the art.


\subsection{Person-Clustering}
\label{sub:res_personclustering}

\noindent 
\textbf{Baselines.} 
To evaluate person-clustering, we compare to two strong baselines stemming from  the best existing face-clustering algorithm, C1C~\cite{Kalogeiton20}. 
The first, B-ReID, is inspired by person Re-ID~\cite{zheng2015scalable,zheng2017unlabeled,li2014deepreid} and uses C1C to cluster body rather than face features. It ignores person-tracks without bodies ({<}2\% of person-tracks). 
For the second, B-C1C, we use regular C1C to cluster faces, with the addition of Stage 3 of \mnameMM for clustering face-less bodies. 

\noindent 
\textbf{Results and analysis.}
Table~\ref{tab:person_results} reports person-clustering results when testing on \dnameemphasis. 
%
For all metrics, \mnameMM (full method) significantly outperforms the strongest baseline by on average 6.1\% in WCP and 11.8\% in NMI.
B-ReID is poor due to frequent clothing changes. \mnameMM outperforms B-C1C 
thanks to (1) the NN distance threshold that prevents incorrect merges and subsequent low-precision clusters, and (2) the multi-modal bridges that merge clusters which face alone cannot. 
\new{This validates that using all available video cues, such as multi-modality and editing structure aids video person-clustering substantially.}
%
\new{The clustering process for a character in \dnameemphasis is visualised qualitatively and quantitatively in Figure~\ref{fig:cluster_res}. \mnameMM improves most upon the baselines on the more unconstrained program sets with many secondary characters and long-tailed character distributions (\eg TBBT, Buffy, Friends, Sherlock). Here, \mnameMM uses the NN distance threshold to keep the clusters of the many characters separated, and then merges any repeated clusters of main-characters via talking person-tracks. The \mnameMM clustering process is visualised in Figure~\ref{fig:qual_clust}.}




\begin{figure*}[th!]
\vspace{-2mm}
     \centering
     \begin{subfigure}[b]{0.18\textwidth}
         \centering
         \includegraphics[width=\textwidth]{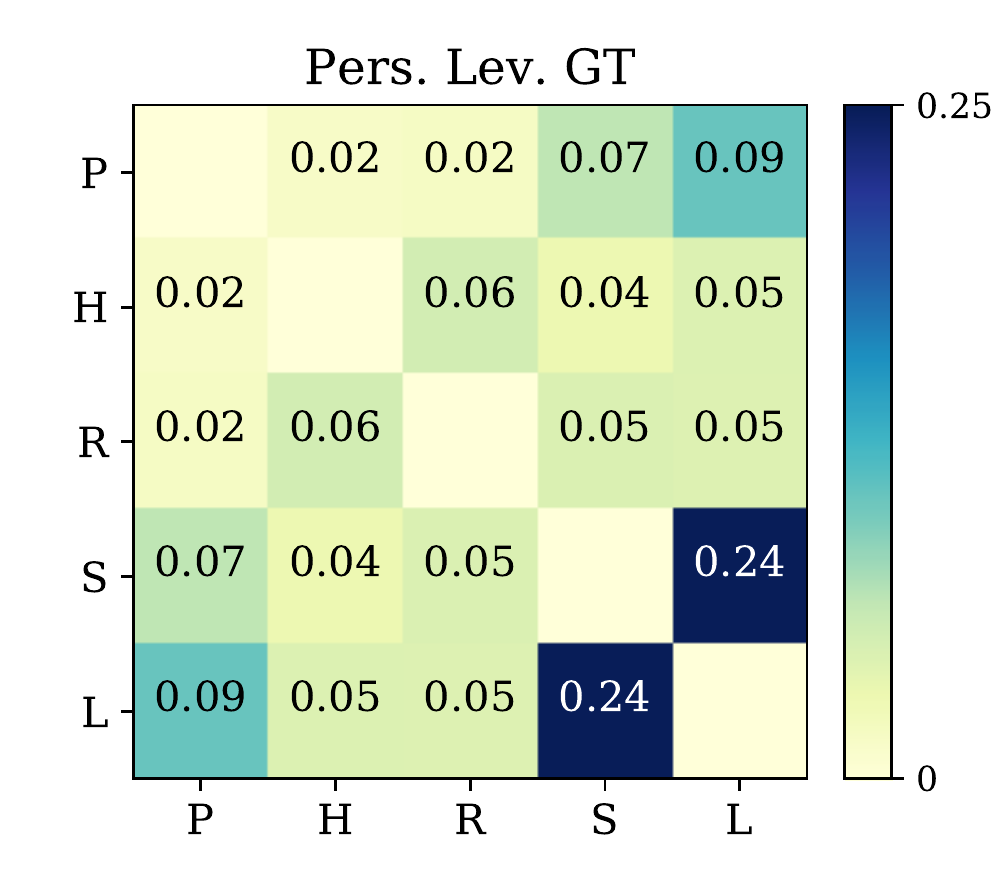}
        \vspace{-7mm}
         \caption{\footnotesize{}}
         \label{SU_GT}
         
     \end{subfigure}
     \hfill
     \begin{subfigure}[b]{0.18\textwidth}
         \centering
         \includegraphics[width=\textwidth]{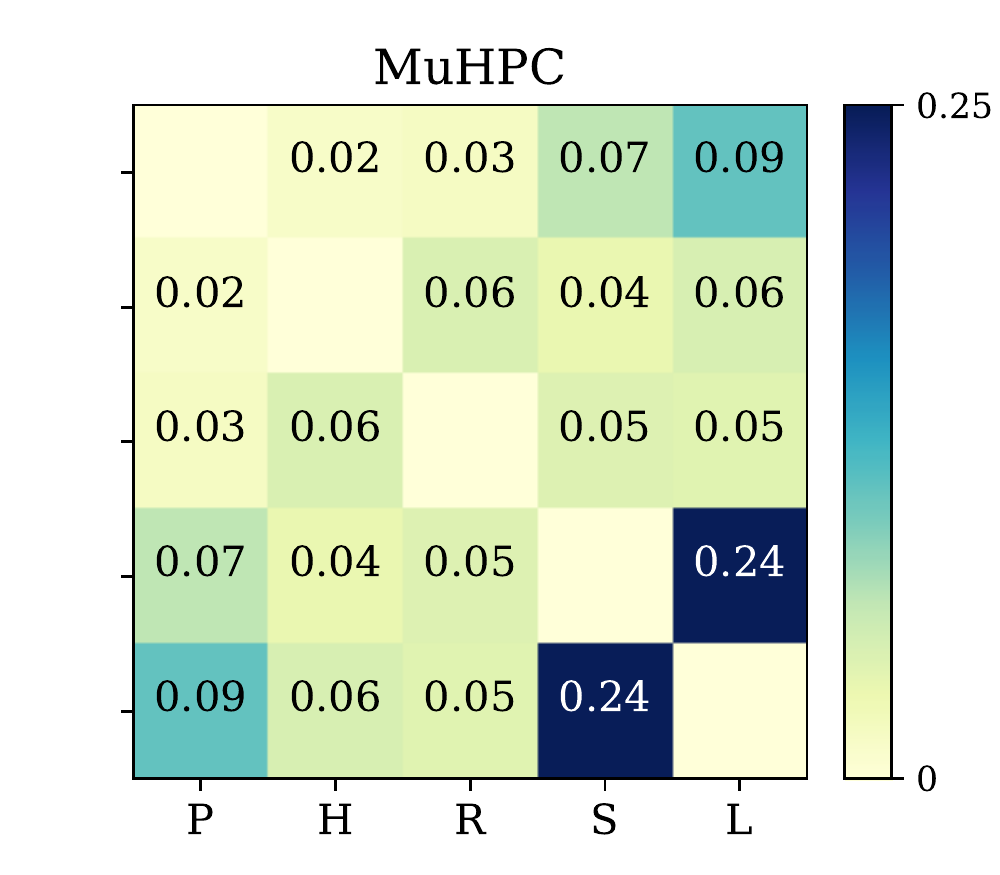}
         \vspace{-7mm}
         \caption{\footnotesize{}}
         \label{SU_MMHPC_absolute}
     \end{subfigure}
     \hfill
     \begin{subfigure}[b]{0.18\textwidth}
         \centering
         \includegraphics[width=\textwidth]{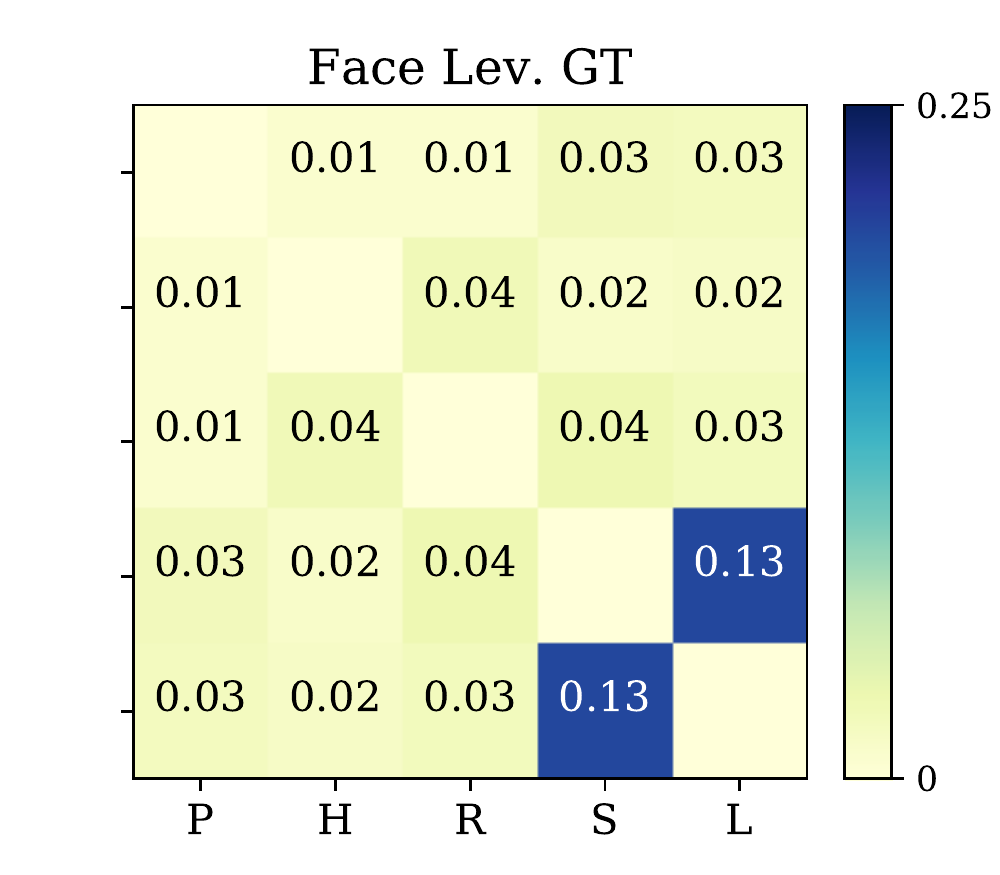}
         \vspace{-7mm}
         \caption{\footnotesize{}}
         \label{SU_Face_absolute}
    \end{subfigure}
    \hfill
    \begin{subfigure}[b]{0.18\textwidth}
         \centering
         \includegraphics[width=\textwidth]{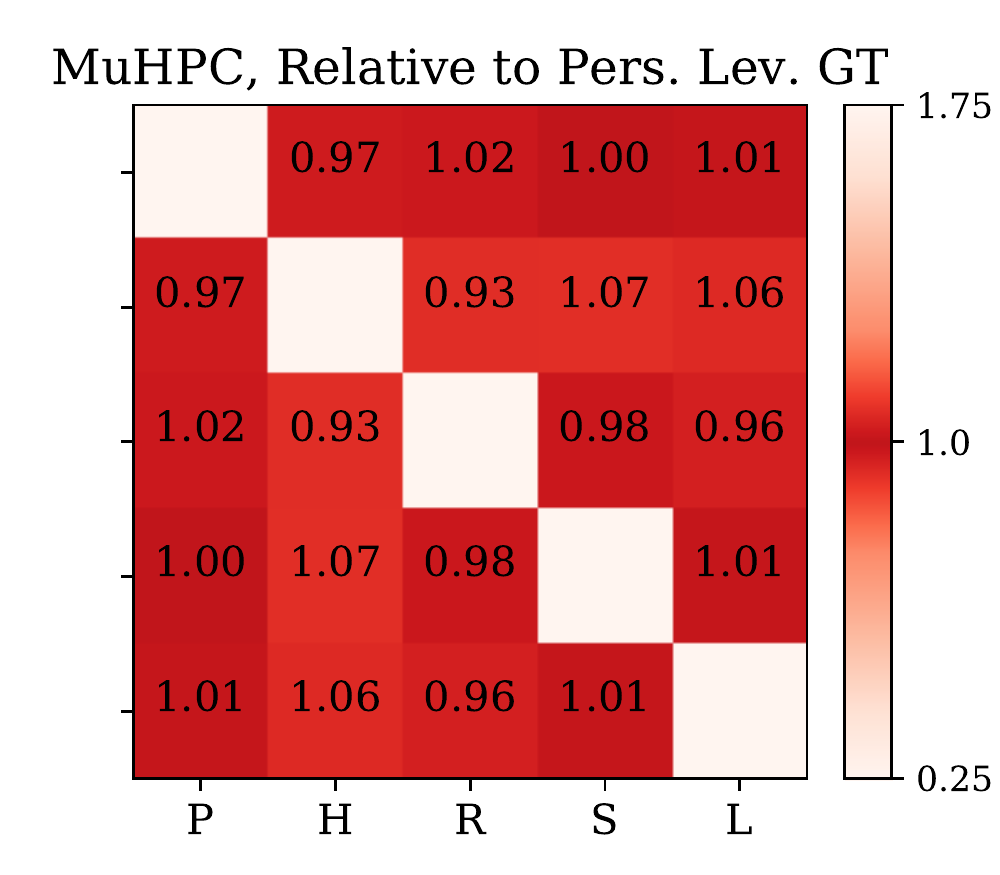}
         \vspace{-7mm}
         \caption{\footnotesize{}}
        \label{SU_MMHPC_relative}
        
    \end{subfigure}
    \hfill
     \begin{subfigure}[b]{0.18\textwidth}
         \centering
         \includegraphics[width=\textwidth]{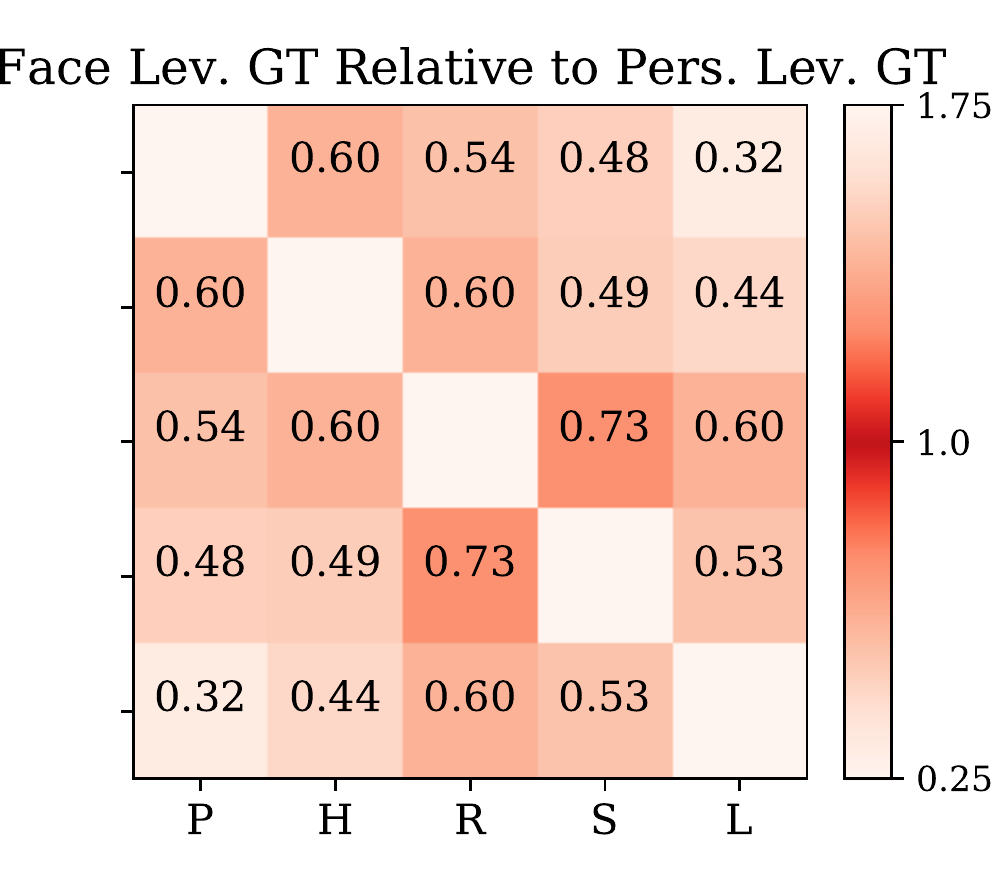}
         \vspace{-7mm}
         \caption{\footnotesize{}}
         \label{SU_Face_relative}
     \end{subfigure}
     \vspace{-3mm}
\caption{\footnotesize{\textbf{ Character co-occurrences for story understanding} between the 5 main characters in the 6 episodes of The Big Bang Theory in \dnameemphasis. (a) The ground truth co-occurrences as a proportion of the 6 videos. It is generated from  \dnameemphasis which annotates each character whenever they are visible. Higher indicates more co-occurrence. (b,c): \mnameMM and face-level clustering co-occurrences; (d,e): \mnameMM and face-level co-occurrences, relative to the ground truth (a). 1.0 indicates that the prediction is the same as the ground truth. Key: P: Penny, H: Howard, R: Raj, S: Sheldon, L:Leonard.}
}
        \label{fig:story_understanding}
\end{figure*}

\subsection{Ablation}    
\label{sub:res_ablation} 
Here, we perform ablations on the different modalities in \mnameMM. Detailed results and parameter sweeps can be found in the appendix. 
Table~\ref{tab:person_results} includes an ablation of the multi-modality, \new{\ie using voice ($\mnameMM_v$ -- Stage 2) or body ($\mnameMM_b$ -- Stage 3) modalities or both (\mnameMM). Experiments without the body modality do not use Stage 3, and instead cluster each face-less body to the temporally-closest (Temporal-NN) body with a face in a nearby shot. Due to the threading structure~\cite{Hoai14e} of edited videos, there is a strong prior that the Temporal-NN is correct.} 

\new{Adding either the voice or body offers a benefit over \mnameMM--, due to the increased discriminative capabilities from an additional modality. $\mnameMM_b$ outperforms $\mnameMM_v$, as there are many face-less bodies in \dnameemphasis, and the body modality allows for these to be clustered correctly. Using the voice in conjunction with the body (\mnameMM) performs best, as their benefits are compounded, and the multi-modal bridges connect clusters with higher purity.
The voice gives a higher boost when used alongside the body modality, as otherwise the multi-modal bridges are merging lower precision clusters.
The voice adds significant benefit in NMI on multiple program-sets. This is impressive as the tight voice thresholds were found \textit{automatically}. Sometimes the voice does not lead to an improvement, due to the absence of speaking person-tracks in merge-able clusters (\eg TBBT). Additionally, the body offers little improvement in the two movies (Hidden Figures, About Last Night) that have many dark scenes and non-distinctive clothing. Here temporal-NN is able to assign face-less bodies to clusters well. Note, NMI increases more than WCP when adding the voice modality, because bridging two high-precision clusters will not greatly effect the purity; however, it leads to increased NMI as there is less identity overlap between the resulting clusters. } 





\subsection{Face-Clustering}
\label{sub:res_faceclustering}

Here, we compare to previous works by experimenting only on face-tracks, excluding person-tracks without faces. 
We compare to FINCH~\cite{finch} (evaluated at the required number of clusters, from~\cite{Kalogeiton20}), BCL~\cite{ball} and C1C~\cite{Kalogeiton20}. \new{For TBBT and Buffy, the face annotations are the same as~\cite{ball,Kalogeiton20}. Here, we do not compare to works that use the less challenging~\cite{Kalogeiton20} subset of the annotations~\cite{sharma2020clusteringsota,rothlingshofer2019self}.} For our method, we present: 
(i) \mnameMM--  uses only face-tracks, \ie exactly the \emph{same} information and features as other methods, hence results are directly comparable; and 
(ii) $\mnameMM_v$ uses face-tracks with multi-modal bridges (\ie voice).
\new{Following}~\cite{Kalogeiton20,ball}, performance is evaluated at frame level. 

Table~\ref{tab:face_results} reports face-clustering results. 
For both \new{AT and OC} protocols, \mnameMM-- significantly outperforms the state of the art in all metrics, as it avoids incorrect merges, hence maintaining cluster purity. 
For instance, NMI, CP and CR boost by +10-14\% for Buffy and TBBT for OC, and by over $10\%$ for WCP averaged across all datasets for AT. 
$\mnameMM_v$ also leads to a boost over \mnameMM-- in most datasets. We observe that the more challenging the dataset, the higher the boosts by multi-modality, \eg +3.8\% in CR for Friends and +7.4\% in NMI for Sherlock. We note that on NMI, WCP, the performance on TBBT is now almost saturated.
A full discussion of results is given in the appendix. 


\begin{table}[t!]
\centering
{\setlength\tabcolsep{5pt}  
\resizebox{\linewidth}{!}{
\begin{tabular}{lc|rrrrcr rrrcr}
\toprule

\multicolumn{1}{l}{\multirow{2}{*}{Method}} & \multirow{2}{*}{protocol} & \multicolumn{4}{c}{\cellcolor{lavenderblue}{\textbf{TBBT}}}  & \cellcolor{lavenderblue}{\textbf{$\#C_s$ = 130}}  &        & \multicolumn{4}{c}{\cellcolor{lavenderblue}{\textbf{Buffy}}}  &  \cellcolor{lavenderblue}{\textbf{$\#C_s$ = 165}}  \\
\multicolumn{1}{c}{}                        &                           & WCP   & NMI   & CP    & CR    & $\#C_p$ & & WCP   & NMI   & CP    & CR    & $\#C_p$  \\ \hline

BCL~\cite{ball}                                        & AT                        & 90.8 & 85.7 &   -    &    -   & 83  & & 85.0 & 78.8 &  -     &    -   & 121    \\
C1C~\cite{Kalogeiton20}                                          & AT                        &   89.2    &  87.4     &  29.1     &  40.9     &   41   &      &   66.3    &  68.8     & 14.9      &    27.1  &  40       \\  \midrule 
\mnameMM--                                      & AT                        & \textbf{99.4}  & \textbf{97.8}    & \textbf{87.8} & \textbf{88.6} & 168 && \textbf{96.1}  & 92.8  & 85.6 & \textbf{85.5} & 223   \\
$\mnameMM_v$                                     & AT                        & \textbf{99.4}  & \textbf{97.8}    & \textbf{87.8} & \textbf{88.6} & 168 & &  \textbf{96.1}  &  \textbf{93.7}  &  \textbf{85.9} & 84.8 & 221 \\ 
\midrule \midrule
Finch~\cite{finch}                                       & OC                        & 90.8  &  80.5     &   46.1    &    44.2   &  \multirow{5}{*}{}&  & 82.9  &   75.3    &   49.6    &   41.0    &    \\
BCL~\cite{ball}                                         & OC                        & 94.0    & 85.0    &   -    &      - &  &  & 86.5  & 77.6  &  -     &  -     &   \\
C1C~\cite{Kalogeiton20}                                         & OC                        & 95.3  & 84.5  &   54.9    &  57.3   &    & & 88.1  & 79.1  &    58.1   &  55.4    &    \\  \midrule
\mnameMM--                                      & OC                        & \textbf{99.1}  & \textbf{97.4}  & \textbf{79.3}  & \textbf{83.0}  &   & & \textbf{95.6}  &  92.2  & \textbf{72.3}  & \textbf{73.8}  &     \\
$\mnameMM_v$                          & OC           &   \textbf{99.1}    &  \textbf{97.4}     &   \textbf{79.3}    & \textbf{83.0}  &  &   &  \textbf{95.6}       &  \textbf{93.1}       &    71.5   &    73.2   &     \\ \midrule
 

\multicolumn{1}{l}{\multirow{1}{*}{ }} & \multirow{1}{*}{ } &  \multicolumn{4}{c}{\cellcolor{lavenderblue}{\textbf{Friends}}}   & \cellcolor{lavenderblue}{\textbf{$\#C_s$ = 239}}  &    & \multicolumn{4}{c}{\cellcolor{lavenderblue}{\textbf{Sherlock}}} &  \cellcolor{lavenderblue}{\textbf{$\#C_s$ = 50}} \\ \midrule

C1C~\cite{Kalogeiton20}                                          & AT                        &  88.2  &  89.8     &  62.4     &     73.2  & 185 & &  76.3&  50.3    & 20.2     & 41.0     &     25      \\ \midrule 
\mnameMM--                                      & AT                        &  \textbf{98.7} & 94.9 & 98.1 & 94.0 & 543 & & \textbf{86.7}   & 60.3     &79.1     & 71.2     & 96     \\
$\mnameMM_v$                                     & AT                        &  98.4 & \textbf{95.9} & 97.7 & \textbf{95.3} & 522 & &  86.3   & \textbf{66.0}     & \textbf{78.4}     & \textbf{74.5}     &     86\\ 
\midrule \midrule
Finch~\cite{finch}                                       & OC  &                       92.2   &  89.9     &     85.2  &    85.6   & \multirow{5}{*}{} & &  81.6 &  58.6    &  \textbf{59.8}    &     56.8 & \multirow{5}{*}{}  \\
C1C~\cite{Kalogeiton20}                                         & OC                        & 94.3 & 93.2 & 79.1  & 85.5 &   & &  81.6 & 53.8 &     40.5 & 51.7     &    \\  \midrule 
\mnameMM--                                      & OC                        &  96.3 & 92.7 & 89.0  & 88.8  & &    & 84.0 & 56.5 & 55.4 & 59.9 &    \\
$\mnameMM_v$                           & OC           &   \textbf{97.1}     & \textbf{94.6}      &    \textbf{92.3}   & \textbf{92.6}      &  &   &    \textbf{85.1}  &     \textbf{63.9} & 59.6     &  \textbf{62.9}    &    \\ \bottomrule

\end{tabular}
}
}
\vspace{-1mm}
\caption{\footnotesize{
\textbf{Face-Clustering Results.} Comparisons to previous state of the art on four program sets, using only face-tracks with unknown (AT), and known (OC) number of clusters. We report metrics averaged over each episode in each program set, and the number of predicted clusters, summed over each episode  ($\#C_p$). 
\mnameMM-- uses only face, 
whereas $\mnameMM_v$ uses the multi-modal bridges from voice and face. Where not reported in respective publications, numbers are computed using official implementations. Finch has no stopping criterion so results for AT are not reported. }}
\vspace{-4mm}
\label{tab:face_results}
\end{table}

\subsection{Enabling Story Understanding}
\label{sub:res_story_understanding}

Here, we explore how close we have come to enabling story understanding. 
Clustering 
people (rather than faces) indicates who is present in a scene (Figure~\ref{fig:teaser_2}) -- an essential and necessary step for predicting character co-occurrences, and hence their interactions~\cite{kukleva2020learning,Marin2020pami} that make up a story. 
Specifically, we ask two questions: 
(1) Can clustering on the face-level predict the co-occurrence of two characters correctly? 
(2) How close is \mnameMM to correctly predicting co-occurrences? To answer these, we experiment with the five main characters from the six episodes of TBBT in \dnameemphasis. Figure~\ref{SU_GT} shows the ground truth (Pers. Lev. GT) co-occurrence of character pairs as a proportion of all frames in the show, 
and hence is a measure of their interaction, \eg Sheldon and Leonard co-occur for 24\% of all frames.

For the first question, we visualise the co-occurrences of characters according to the face-track annotations (Face Lev. GT) in \dnameemphasis (shown in absolute terms in Figure~\ref{SU_Face_absolute}, and relative to the GT in Figure~\ref{SU_Face_relative}). 
The face-level co-occurrences are poor -- with an average error from GT of 48\%. For instance, Penny and Leonard, whose romance is a main story-line, are shown to co-occur in only 3\% of the videos vs the GT 9\%. Furthermore, the GT shows that this is the second most commonly occurring pair; nevertheless, the face-level annotations fail to pick up that it is significant relative to other pairs. This is expected as often one or more characters do not show their face when appearing together (Figure~\ref{fig:teaser_2}). Hence, any co-occurrences predicted from the face-level are a limited foundation for story understanding.

For the second question, we cluster with \mnameMM and assign each cluster to the character that appears most within it (Figures~\ref{SU_MMHPC_absolute},~\ref{SU_MMHPC_relative}). 
We observe that these predictions are very close to the GT, with an average error of just 3\% (Figure~\ref{SU_MMHPC_relative}). This impressive result shows that the presence of each character, their co-occurrence and hence their possible interactions can be found completely automatically and accurately using our proposed method. This provides a rich and informative foundation for story understanding.





\ifx 
First, the nearest neighbour distance threshold,  $\tau_f^{\text{tight}}$. A high value allows for incorrect merging of clusters of different identities that are far apart. A value too low which leads to a strict stopping criterion that stops merging clusters too early. This is shown in Figure~\ref{fig:cluster_example} (right). The value is chosen as a balance between these two factors. 
Similarly for the multi-modality bridging thresholds, $\tau_f^{\text{loose}}$, $\tau_v^{\text{tight}}$,  $\tau_b^{\text{tight}}$, lower values indicate a stricter threshold on the distance between tracks. We find that for high values for $\tau_v^{\text{tight}}$ and $\tau_b^{\text{tight}}$, many false positive bridges are made and clustering quality decreases. This is because for the less discriminative modalities of body and voice, a low similarity cannot be taken as a confident indicator of two tracks being from the same identity. By keeping  $\tau_v^{\text{tight}}$ and $\tau_b^{\text{tight}}$ low (0.4 and 0.35, respectively) and raising the face threshold to $\tau_f^{\text{loose}} = 0.55$, the optimal balance is struck, where both modalities confidently agree to bridges between clusters that the face modality alone could not bridge. Full quantitative analysis can be found in the supplementary material.
\fi 




\ifx
\subsection{Discussion}
\label{sub:res_dicussion}
\mnameMM shows significant boosts over all baselines, ablations and previous methods for both face-clustering and person-clustering. 
Furthermore, using extra modalities leads to a consistent advantage. 
The person-clustering results (Table~\ref{tab:person_results}) show that there is room for improvement. 

\noindent \textbf{Metric assessment.}
Interestingly, for both AT and OC protocols, the WCP results are much higher than other metrics (\eg +8-17\% in Table~\ref{tab:person_results}). 
This shows that WCP \emph{alone} is not suitable for measuring the clustering performance, even when evaluating on the oracle number of clusters (OC), thus highlighting the importance of our additional CP and CR metrics. 
For AT, when predicting more clusters (est \#C is higher than the oracle \#C) high purity is expected. 
Notably, in all cases NMI also increases, thus indicating the high clustering quality, showing that there is far less class intersection between clusters for our method.
 \fi

\section{Conclusions}
\label{sec:conclusions}
In this work we propose \mnameMM, a novel method for multi-modal person-clustering in videos.  For evaluation we introduced \dnameemphasis, \new{the largest and most diverse dataset of its kind. We showed that using 
all available video cues is essential for person-clustering, leading to significant improvements on  \dnameemphasis, and to state-of-the-art performance for face-clustering. Importantly, we demonstrated that \mnameMM allows each character appearance and co-occurrence to be predicted completely automatically and accurately. We hope this can support downstream story understanding tasks such as learning interactions and relationships~\cite{kukleva2020learning}.}

\section*{Acknowledgement}
\noindent
\small{We are grateful to Bruno Korbar and Sagar Vaze for their helpful comments, and to Maya Gulieva, Shaya Ghadimi, and DK for their help with annotation. AB is funded by an EPSRC DTA Studentship. This work is supported by the EPSRC Programme Grants Seebibyte EP/M013774/1 and 
VisualAI EP/T028572/1, and by a CNRS INS2I 2020 grant.}

{\small
\bibliographystyle{ieee_fullname}
\bibliography{shortstrings,arxiv_biblio}

\begin{thebibliography}{10}\itemsep=-1pt

\bibitem{asano2020labelling}
Yuki Asano, Mandela Patrick, Christian Rupprecht, and Andrea Vedaldi.
\newblock Labelling unlabelled videos from scratch with multi-modal
  self-supervision.
\newblock {\em NeurIPS}, 2020.

\bibitem{Awad2019TRECVID2A}
G. Awad, A. Butt, K. Curtis, Y. Lee, J. Fiscus, A. Godil, A. Delgado, Jesse
  Zhang, Eliot Godard, Lukas~L. Diduch, A.~F. Smeaton, Yyette Graham, Wessel
  Kraaij, and Georges Qu{\'e}not.
\newblock Trecvid 2019: An evaluation campaign to benchmark video activity
  detection, video captioning and matching, and video search \& retrieval.
\newblock {\em ArXiv}, abs/2009.09984, 2019.

\bibitem{Bain20}
Max Bain, Arsha Nagrani, Andrew Brown, and Andrew Zisserman.
\newblock Condensed movies: Story based retrieval with contextual embeddings.
\newblock In {\em ACCV}, 2020.

\bibitem{bauml2013semi}
Martin Bauml, Makarand Tapaswi, and Rainer Stiefelhagen.
\newblock Semi-supervised learning with constraints for person identification
  in multimedia data.
\newblock In {\em Proc. CVPR}, 2013.

\bibitem{berg2004cvpr}
Tamara~L Berg, Alexander~C Berg, Jaety Edwards, Michael Maire, Ryan White,
  Yee-Whye Teh, Erik Learned-Miller, and David~A Forsyth.
\newblock Names and faces in the news.
\newblock In {\em Proc. CVPR}, 2004.

\bibitem{Bojanowski13}
P. Bojanowski, F. Bach, , I. Laptev, J. Ponce, C. Schmid, and J. Sivic.
\newblock Finding actors and actions in movies.
\newblock In {\em Proc. ICCV}, 2013.

\bibitem{Brown21}
Andrew Brown, Ernesto Coto, and Andrew Zisserman.
\newblock Automated video labelling: Identifying faces by corroborative
  evidence.
\newblock In {\em MIPR}, 2021.

\bibitem{Brown20}
Andrew Brown, Weidi Xie, Vicky Kalogeiton, and Andrew Zisserman.
\newblock Smooth-{AP}: Smoothing the path towards large-scale image retrieval.
\newblock In {\em European Conference on Computer Vision (ECCV)}, 2020.

\bibitem{cai2018cascade}
Zhaowei Cai and Nuno Vasconcelos.
\newblock Cascade r-cnn: Delving into high quality object detection.
\newblock In {\em Proc. CVPR}, 2018.

\bibitem{Cao18}
Qiong Cao, Li Shen, Weidi Xie, Omkar~M. Parkhi, and Andrew Zisserman.
\newblock {VGGFace2}: A dataset for recognising faces across pose and age.
\newblock In {\em Proc. Int. Conf. Autom. Face and Gesture Recog.}, 2018.

\bibitem{chung2020in}
Joon~Son Chung, Jaesung Huh, Seongkyu Mun, Minjae Lee, Hee~Soo Heo, Soyeon
  Choe, Chiheon Ham, Sunghwan Jung, Bong-Jin Lee, and Icksang Han.
\newblock In defence of metric learning for speaker recognition.
\newblock In {\em INTERSPEECH}, 2020.

\bibitem{Chung18a}
Joon~Son Chung, Arsha Nagrani, and Andrew Zisserman.
\newblock {VoxCeleb2}: Deep speaker recognition.
\newblock In {\em INTERSPEECH}, 2018.

\bibitem{cinbis2011iccv}
Ramazan~Gokberk Cinbis, Jakob Verbeek, and Cordelia Schmid.
\newblock Unsupervised metric learning for face identification in tv video.
\newblock In {\em Proc. ICCV}, 2011.

\bibitem{cour2009learning}
Timothee Cour, Benjamin Sapp, Chris Jordan, and Ben Taskar.
\newblock Learning from ambiguously labeled images.
\newblock In {\em Proc. CVPR}, 2009.

\bibitem{Cour10}
T. Cour, B. Sapp, A. Nagle, and B. Taskar.
\newblock Talking pictures: Temporal grouping and dialog-supervised person
  recognition.
\newblock In {\em Proc. CVPR}, 2010.

\bibitem{de2012constrained}
Renato~Cordeiro de Amorim.
\newblock Constrained clustering with minkowski weighted k-means.
\newblock In {\em CINTI}, 2012.

\bibitem{ercolessi2012stoviz}
Philippe Ercolessi, Herv{\'e} Bredin, and Christine S{\'e}nac.
\newblock Stoviz: story visualization of tv series.
\newblock In {\em Proc. ACMMM}, 2012.

\bibitem{Everingham06a}
Mark Everingham, Josef Sivic, and Andrew Zisserman.
\newblock ``{H}ello! {M}y name is... {Buffy}'' -- automatic naming of
  characters in {TV} video.
\newblock In {\em Proc. BMVC}, 2006.

\bibitem{Everingham09}
Mark Everingham, Josef Sivic, and Andrew Zisserman.
\newblock Taking the bite out of automatic naming of characters in {TV} video.
\newblock {\em Image and Vision Computing}, 2009.

\bibitem{Fitzgibbon02}
Andrew~W. Fitzgibbon and Andrew Zisserman.
\newblock On affine invariant clustering and automatic cast listing in movies.
\newblock In {\em Proc. ECCV}, 2002.

\bibitem{ghaleb2015accio}
Esam Ghaleb, Makarand Tapaswi, Ziad Al-Halah, Hazim~Kemal Ekenel, and Rainer
  Stiefelhagen.
\newblock Accio: A data set for face track retrieval in movies across age.
\newblock In {\em Proc. ICMR}, 2015.

\bibitem{isthatyou}
Matthieu Guillaumin, Jakob Verbeek, and Cordelia Schmid.
\newblock Is that you? metric learning approaches for face identification.
\newblock 2009.

\bibitem{guo2016ms}
Yandong Guo, Lei Zhang, Yuxiao Hu, Xiaodong He, and Jianfeng Gao.
\newblock {MS-Celeb-1M}: A dataset and benchmark for large-scale face
  recognition.
\newblock In {\em ECCV}, 2016.

\bibitem{haurilet2016naming}
Monica-Laura Haurilet, Makarand Tapaswi, Ziad Al-Halah, and Rainer
  Stiefelhagen.
\newblock Naming tv characters by watching and analyzing dialogs.
\newblock In {\em Proc. WACV}, 2016.

\bibitem{He16}
Kaiming He, Xiangyu Zhang, Shaoqing Ren, and Jian Sun.
\newblock Deep residual learning for image recognition.
\newblock In {\em Proc. CVPR}, 2016.

\bibitem{he2016deep}
Kaiming He, Xiangyu Zhang, Shaoqing Ren, and Jian Sun.
\newblock Deep residual learning for image recognition.
\newblock In {\em Proc. CVPR}, 2016.

\bibitem{he2018aaai}
Yue He, Kaidi Cao, Cheng Li, and Chen~Change Loy.
\newblock Merge or not? learning to group faces via imitation learning.
\newblock In {\em AAAI}, 2018.

\bibitem{ho2003cvpr}
Jeffrey Ho, Ming-Husang Yang, Jongwoo Lim, Kuang-Chih Lee, and David Kriegman.
\newblock Clustering appearances of objects under varying illumination
  conditions.
\newblock In {\em Proc. CVPR}, 2003.

\bibitem{Hoai14e}
Minh Hoai and Andrew Zisserman.
\newblock Thread-safe: Towards recognizing human actions across shot
  boundaries.
\newblock In {\em Asian Conference on Computer Vision}, 2014.

\bibitem{hu2018squeeze}
Jie Hu, Li Shen, and Gang Sun.
\newblock Squeeze-and-excitation networks.
\newblock In {\em Proc. CVPR}, 2018.

\bibitem{huang2018person}
Qingqiu Huang, Wentao Liu, and Dahua Lin.
\newblock Person search in videos with one portrait through visual and temporal
  links.
\newblock In {\em Proc. ECCV}, 2018.

\bibitem{huang2020movienet}
Qingqiu Huang, Yu Xiong, Anyi Rao, Jiaze Wang, and Dahua Lin.
\newblock Movienet: A holistic dataset for movie understanding.
\newblock In {\em Proc. ECCV}, 2020.

\bibitem{JarvisP73}
Raymond~Austin Jarvis and Edward~A. Patrick.
\newblock Clustering using a similarity measure based on shared near neighbors.
\newblock {\em {IEEE} Trans. Computers}, 1973.

\bibitem{jin2017iccv}
SouYoung Jin, Hang Su, Chris Stauffer, and Erik Learned-Miller.
\newblock End-to-end face detection and cast grouping in movies using
  erdos-renyi clustering.
\newblock In {\em Proc. ICCV}, 2017.

\bibitem{joon2015person}
Seong Joon~Oh, Rodrigo Benenson, Mario Fritz, and Bernt Schiele.
\newblock Person recognition in personal photo collections.
\newblock In {\em Proc. ICCV}, 2015.

\bibitem{Kalogeiton20}
Vicky Kalogeiton and Andrew Zisserman.
\newblock Constrained video face clustering using 1nn relations.
\newblock In {\em Proc. BMVC}, 2020.

\bibitem{Kostinger11}
M. K{\"o}stinger, P. Wohlhart, P. Roth, and H. Bischof.
\newblock Learning to recognize faces from videos and weakly relatedinformation
  cues.
\newblock In {\em AVSS}, 2011.

\bibitem{kuhn1955hungarian}
Harold~W Kuhn.
\newblock The hungarian method for the assignment problem.
\newblock {\em Naval research logistics quarterly}, 1955.

\bibitem{kukleva2020learning}
Anna Kukleva, Makarand Tapaswi, and Ivan Laptev.
\newblock Learning interactions and relationships between movie characters.
\newblock In {\em Proc. CVPR}, 2020.

\bibitem{li2014deepreid}
Wei Li, Rui Zhao, Tong Xiao, and Xiaogang Wang.
\newblock Deepreid: Deep filter pairing neural network for person
  re-identification.
\newblock In {\em Proc. CVPR}, 2014.

\bibitem{lin2018cvpr}
Wei-An Lin, Jun-Cheng Chen, Carlos~D Castillo, and Rama Chellappa.
\newblock Deep density clustering of unconstrained faces.
\newblock In {\em Proc. CVPR}, 2018.

\bibitem{lowe2004distinctive}
David~G Lowe.
\newblock Distinctive image features from scale-invariant keypoints.
\newblock {\em International journal of computer vision}, 2004.

\bibitem{manning2008introduction}
Christopher~D Manning, Prabhakar Raghavan, and Hinrich Sch{\"u}tze.
\newblock {\em Introduction to information retrieval}.
\newblock Cambridge university press, 2008.

\bibitem{Marin19a}
M.~J. Marin-Jimenez, V. Kalogeiton, P. Medina-Suarez, and A. Zisserman.
\newblock {LAEO-Net}: revisiting people {Looking At Each Other} in videos.
\newblock In {\em Proc. CVPR}, 2019.

\bibitem{Marin2020pami}
M.~J. Marin-Jimenez, V. Kalogeiton, P. Medina-Suarez, and A. Zisserman.
\newblock {LAEO-Net++}: revisiting people {Looking At Each Other} in videos.
\newblock In {\em IEEE PAMI}, 2020.

\bibitem{Nagrani18c}
Arsha Nagrani, Samuel Albanie, and Andrew Zisserman.
\newblock Learnable pins: Cross-modal embeddings for person identity.
\newblock In {\em Proc. ECCV}, 2018.

\bibitem{Nagrani18a}
Arsha Nagrani, Samuel Albanie, and Andrew Zisserman.
\newblock Seeing voices and hearing faces: Cross-modal biometric matching.
\newblock In {\em Proc. CVPR}, 2018.

\bibitem{Nagrani17b}
Arsha Nagrani and Andrew Zisserman.
\newblock From benedict cumberbatch to sherlock holmes: Character
  identification in tv series without a script.
\newblock In {\em Proc. BMVC}, 2017.

\bibitem{otto2017pami}
Charles Otto, Dayong Wang, and Anil~K Jain.
\newblock Clustering millions of faces by identity.
\newblock {\em IEEE PAMI}, 2017.

\bibitem{hannah}
Alexey Ozerov, Jean-Ronan Vigouroux, Louis Chevallier, and Patrick P{\'e}rez.
\newblock On evaluating face tracks in movies.
\newblock In {\em Intl. Conf. Image Proc.}, 2013.

\bibitem{Parkhi15a}
Omkar~M. Parkhi, Esa Rahtu, and Andrew Zisserman.
\newblock It's in the bag: Stronger supervision for automated face labelling.
\newblock In {\em ICCV Workshop: Describing and Understanding Video \& The
  Large Scale Movie Description Challenge}, 2015.

\bibitem{poignant2017multimodal}
Johann Poignant, Herv{\'e} Bredin, and Claude Barras.
\newblock Multimodal person discovery in broadcast tv: lessons learned from
  mediaeval 2015.
\newblock {\em Multimedia Tools and Applications}, 2017.

\bibitem{ramanan2007leveraging}
Deva Ramanan, Simon Baker, and Sham Kakade.
\newblock Leveraging archival video for building face datasets.
\newblock In {\em Proc. ICCV}, 2007.

\bibitem{ramanathan2014linking}
Vignesh Ramanathan, Armand Joulin, Percy Liang, and Li Fei-Fei.
\newblock Linking people in videos with “their” names using coreference
  resolution.
\newblock In {\em Proc. ECCV}, 2014.

\bibitem{rao2020local}
Anyi Rao, Linning Xu, Yu Xiong, Guodong Xu, Qingqiu Huang, Bolei Zhou, and
  Dahua Lin.
\newblock A local-to-global approach to multi-modal movie scene segmentation.
\newblock In {\em Proc. CVPR}, 2020.

\bibitem{Partitional}
Chandan Reddy and Bhanukiran Vinzamuri.
\newblock {\em A Survey of Partitional and Hierarchical Clustering Algorithms}.
\newblock 2018.

\bibitem{rothlingshofer2019self}
Veith Rothlingshofer, Vivek Sharma, and Rainer Stiefelhagen.
\newblock Self-supervised face-grouping on graphs.
\newblock In {\em Proc. ACMMM}, 2019.

\bibitem{TVD}
Anindya Roy, Camille Guinaudeau, Herv{\'e} Bredin, and Claude Barras.
\newblock Tvd: a reproducible and multiply aligned tv series dataset.
\newblock In {\em LREC}, 2014.

\bibitem{finch}
Saquib Sarfraz, Vivek Sharma, and Rainer Stiefelhagen.
\newblock Efficient parameter-free clustering using first neighbor relations.
\newblock In {\em Proc. CVPR}, 2019.

\bibitem{sharma2019self_before34}
Vivek Sharma, Makarand Tapaswi, M~Saquib Sarfraz, and Rainer Stiefelhagen.
\newblock Self-supervised learning of face representations for video face
  clustering.
\newblock In {\em Proc. Int. Conf. Autom. Face and Gesture Recog.}, 2019.

\bibitem{sharma2019video34}
Vivek Sharma, Makarand Tapaswi, M~Saquib Sarfraz, and Rainer Stiefelhagen.
\newblock Video face clustering with self-supervised representation learning.
\newblock {\em IEEE Transactions on Biometrics, Behavior, and Identity
  Science}, 2019.

\bibitem{sharma2020clusteringsota}
Vivek Sharma, Makarand Tapaswi, M~Saquib Sarfraz, and Rainer Stiefelhagen.
\newblock Clustering based contrastive learning for improving face
  representations.
\newblock {\em Proc. Int. Conf. Autom. Face and Gesture Recog.}, 2020.

\bibitem{Thinking_Through}
Kate Sim, Andrew Brown, and Amelia Hassoun.
\newblock Thinking through and writing about research ethics beyond "broader
  impact".
\newblock {\em CoRR}, 2021.

\bibitem{Sivic09}
Josef Sivic, Mark Everingham, and Andrew Zisserman.
\newblock ``{W}ho are you?'' -- learning person specific classifiers from
  video.
\newblock In {\em Proc. CVPR}, 2009.

\bibitem{Sivic06a}
Josef Sivic, C.~Larry Zitnick, and Rick Szeliski.
\newblock Finding people in repeated shots of the same scene.
\newblock In {\em Proc. BMVC}, 2006.

\bibitem{somandepalli2020multi}
Krishna Somandepalli, Rajat Hebbar, and Shrikanth Narayanan.
\newblock Multi-face: Self-supervised multiview adaptation for robust face
  clustering in videos.
\newblock {\em arXiv preprint arXiv:2008.11289}, 2020.

\bibitem{tapaswi2012cvpr}
Makarand Tapaswi, Martin B{\"a}uml, and Rainer Stiefelhagen.
\newblock ``knock! knock! who is it?'' probabilistic person identification in
  tv-series.
\newblock In {\em Proc. CVPR}, 2012.

\bibitem{tapaswi2014storygraphs}
Makarand Tapaswi, Martin Bauml, and Rainer Stiefelhagen.
\newblock Storygraphs: visualizing character interactions as a timeline.
\newblock In {\em Proc. CVPR}, 2014.

\bibitem{ball}
Makarand Tapaswi, Marc~T Law, and Sanja Fidler.
\newblock Video face clustering with unknown number of clusters.
\newblock In {\em Proc. ICCV}, 2019.

\bibitem{tapaswi2014total}
Makarand Tapaswi, Omkar~M Parkhi, Esa Rahtu, Eric Sommerlade, Rainer
  Stiefelhagen, and Andrew Zisserman.
\newblock Total cluster: A person agnostic clustering method for broadcast
  videos.
\newblock In {\em Proc. ICVGIP}, 2014.

\bibitem{van2020learning}
Wouter Van~Gansbeke, Simon Vandenhende, Stamatios Georgoulis, Marc Proesmans,
  and Luc Van~Gool.
\newblock Learning to classify images without labels.
\newblock {\em arXiv preprint arXiv:2005.12320}, 2020.

\bibitem{vpcd_website}
VGG.
\newblock Video person clustering dataset.
\newblock \url{https://www.robots.ox.ac.uk/~vgg/data/Video_Person_Clustering/},
  2021.

\bibitem{vicol2018moviegraphs}
Paul Vicol, Makarand Tapaswi, Lluis Castrejon, and Sanja Fidler.
\newblock Moviegraphs: Towards understanding human-centric situations from
  videos.
\newblock In {\em Proc. CVPR}, 2018.

\bibitem{wagstaff2001icml}
Kiri Wagstaff, Claire Cardie, Seth Rogers, Stefan Schr{\"o}dl, et~al.
\newblock Constrained k-means clustering with background knowledge.
\newblock In {\em Proc. ICML}, 2001.

\bibitem{wei2018person}
Longhui Wei, Shiliang Zhang, Wen Gao, and Qi Tian.
\newblock Person transfer gan to bridge domain gap for person
  re-identification.
\newblock In {\em Proc. CVPR}, 2018.

\bibitem{Wohlhart11}
P. Wohlhart, M. K{\"o}stinger, P.~M. Roth, and H. Bischof.
\newblock Multiple instance boosting for face recognition in videos.
\newblock In {\em DAGM-Symposium}, 2011.

\bibitem{wu2013iccv}
Baoyuan Wu, Siwei Lyu, Bao-Gang Hu, and Qiang Ji.
\newblock Simultaneous clustering and tracklet linking for multi-face tracking
  in videos.
\newblock In {\em Proc. ICCV}, 2013.

\bibitem{wu2013cvpr}
Baoyuan Wu, Yifan Zhang, Bao-Gang Hu, and Qiang Ji.
\newblock Constrained clustering and its application to face clustering in
  videos.
\newblock In {\em Proc. CVPR}, 2013.

\bibitem{xia2020online}
Jiangyue Xia, Anyi Rao, Qingqiu Huang, Linning Xu, Jiangtao Wen, and Dahua Lin.
\newblock Online multi-modal person search in videos.
\newblock In {\em Proc. ECCV}, 2020.

\bibitem{xiao2014eccv}
Shijie Xiao, Mingkui Tan, and Dong Xu.
\newblock Weighted block-sparse low rank representation for face clustering in
  videos.
\newblock In {\em Proc. ECCV}, 2014.

\bibitem{Xie19a}
Weidi Xie, Arsha Nagrani, Joon~Son Chung, and Andrew Zisserman.
\newblock Utterance-level aggregation for speaker recognition in the wild.
\newblock In {\em Proc. ICASSP}, 2019.

\bibitem{zhang2015beyond}
Ning Zhang, Manohar Paluri, Yaniv Taigman, Rob Fergus, and Lubomir Bourdev.
\newblock Beyond frontal faces: Improving person recognition using multiple
  cues.
\newblock In {\em Proc. CVPR}, 2015.

\bibitem{zhang2016eccv}
Zhanpeng Zhang, Ping Luo, Chen~Change Loy, and Xiaoou Tang.
\newblock Joint face representation adaptation and clustering in videos.
\newblock In {\em Proc. ECCV}, 2016.

\bibitem{zheng2015scalable}
Liang Zheng, Liyue Shen, Lu Tian, Shengjin Wang, Jingdong Wang, and Qi Tian.
\newblock Scalable person re-identification: A benchmark.
\newblock In {\em Proc. ICCV}, 2015.

\bibitem{zheng2017unlabeled}
Zhedong Zheng, Liang Zheng, and Yi Yang.
\newblock Unlabeled samples generated by gan improve the person
  re-identification baseline in vitro.
\newblock In {\em Proc. ICCV}, 2017.

\end{thebibliography}
}

\clearpage
\appendix
\section*{\LARGE{Appendix}}




\section{Broader Impact}

\label{sup:broader_imapct}
Video Person-Clustering is an appealing topic in Computer Vision, with many downstream applications such as story understanding, video navigation,  and video organisation. A successful person-clustering framework (such as that presented in this work) takes a significant step towards realising these applications by alleviating the tremendous annotation cost that would otherwise be necessary. 

For all potential impacts and applications of video person-clustering, it is essential that the datasets that methods are evaluated on are representative of the real-world in which they (or their downstream applications) may be deployed~\cite{Thinking_Through}. This is essential if the research is to be accessible by different communities around the world. A representative dataset can accurately foreshadow and ultimately prevent any algorithmic discrimination on specific demographic groups. Previous person-clustering datasets (which focused on the narrower task of face-clustering) were non-representative of most demographic groups. To this end, in this work we presented \dnameemphasis, which represents a wide and diverse range of characters, and so is more representative of the diversity in the real-world. 

The person-clustering task aims at recognising and clustering identities. Re-identifying people in the real-world generally poses a threat to their privacy, and could carry risks if used inappropriately. In \dnameemphasis however, the identities are all actors playing the part of characters. This is not private data, and none of the videos have been obtained from social media or search engines. All videos in \dnameemphasis are in fact from public films and television material.



\section{\dnameemphasis Details}
\label{sup:dataset_details}
Here, we give additional details on the annotation (Section~\ref{sup:sub:data_collection}) and feature extraction (Section~\ref{sup:sub:data_features}) process for the body-tracks in \dnameemphasis. These sections are complementary to Sections  4.2 \& 4.3 in the main manuscript. We then give further statistics and details of the voice-tracks in \dnameemphasis (Section~\ref{voice_track_stats}).

\subsection{Annotation Process}
\label{sup:sub:data_collection}
Here, we provide additional details for the body-track annotation in \dnameemphasis. To set the scene, we have body-tracks computed for all program sets in \dnameemphasis. The task at this stage is to annotate the body-tracks with the names of the characters that are annotated in the face-tracks.

The body-tracks fall into two categories, which are annotated separately.
(1) The body-track shows the person from the front and contains a visible, annotated face. For these cases we automatically label the body-tracks by making assignments to labelled face-tracks. Within each shot, the assignment is done using the Hungarian Algorithm~\cite{kuhn1955hungarian} with a cost function of the spatial intersection over union (IOU) between face and body-tracks in the frames that they co-occur. If there are more body-tracks than face-tracks, then a body-track can not be assigned, and vice-versa. In ~95\% of cases this association is trivial and the assignment proceeds automatically. Where multiple assignment costs for the same face-track are below a threshold, indicating that the assignment was non-trivial, we instead make the assignments manually. 
(2) The body-track does not contain a visible face, \ie the back is turned to the camera. 
We manually annotate all of these cases throughout each video. On average, 10-15\% of body-tracks correspond to manually labelled bodies from behind.

\subsection{Feature Extraction}
\label{sup:sub:data_features}
Here, we describe in more detail the feature extraction process for the body-tracks. 

Features are extracted from each of the body-tracks using a ResNet50 architecture~\cite{he2016deep}. 
Our goal is to train the body features to discriminate identity based on the highly discriminative clothing that people are wearing. 
We train a ResNet50 on the CSM dataset~\cite{huang2018person}, which contains identity-labelled body detections from movies. This dataset contains the same label for all body detections of each identity, regardless of their clothing. Instead, we decompose the samples for each class (identity) in CSM into sub-classes containing images of the same identity in the same outfit. Our assumption is that if two detections occur close-by temporally within the same movie, then the person is likely to be wearing the same clothing. Each body detection is annotated with the shot that the detection is found in. We cluster the body detections in each class according to their temporal location, resulting in several sub-classes for each identity, where they are wearing the same clothing. We train the model in a contrastive manner using the Smooth-AP loss from~\cite{Brown20}. For the network to be variant to both identity and clothing, we sample positives from the same identity wearing the same outfit, and negatives from different identities. 


\subsection{\dnameemphasis Voice-Track Statistics}
\label{voice_track_stats}
\begin{figure}[t!]
\begin{center}
\includegraphics[width=0.95\linewidth]{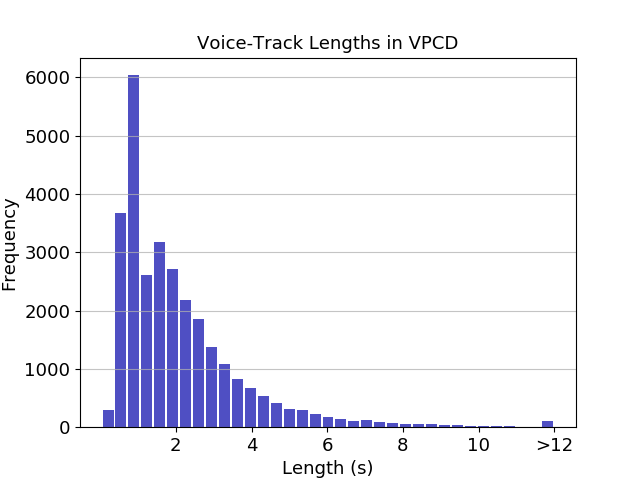}
\end{center}
\vspace{-5mm}
  \caption{\footnotesize{\textbf{Voice-track lengths in \dnameemphasis.} The distribution of all voice-track lengths in \dnameemphasis.}
  }
  \vspace{-2mm}
\label{fig:voice_lengths}
\end{figure}

Here, we give further details and statistics for the voice-tracks in \dnameemphasis. In total, there are 27,163 voice-tracks in \dnameemphasis (Table~\ref{tab:voice_tracks}). This includes annotations for the `laughter' track from the live studio audience in TBBT and Friends, and additionally laughter from each character in all program sets. Features, and the associated annotations for all of these voice-tracks are provided for future research use with \dnameemphasis. The distribution of lengths of these voice-tracks is shown in Figure~\ref{fig:voice_lengths}. These figures for the number of voice-tracks are different to those provided in Table 1 in the main manuscript. \mnameMM implements a pre-processing step on the voice-tracks, such that only the most identity-discriminating voice-tracks are used in the clustering process (explained in Section~\ref{sup:implement_details}).

\section{Implementation Details}
\label{sup:implement_details}
In this section, we give details on a pre-processing step for \mnameMM, which aims to remove voice-tracks that might not be identity-discriminating from the clustering process. Some of the voice-tracks in  \mnameMM are not used, due to overlap between multiple voice-tracks, or due to them being too short. Here, we explain this process, and provide statistics on how many voice-tracks are ignored at this stage (Table~\ref{tab:voice_tracks}). 
First, the temporal overlap between multiple voice-tracks. Our goal here is to use the voice-track features as a discriminative signal for identity. If multiple voice-tracks from different identities have large temporal overlap, then the resulting features will be very similar, and they will not provide a good identity-discriminating signal. We choose to ignore any voice-tracks that have 20\% overlap with a different voice-track. 
Second, the temporal length of the voice-tracks.  As shown in~\cite{Xie19a}, there is a strong positive correlation between the discriminative capabilities of voice-track features and the length of the voice-track. In order to maximise the discriminativeness of the voice-track features, we ignore those that are less than 1 second in length. Table~\ref{tab:voice_tracks} shows the total number of voice-track annotations in  \dnameemphasis before (``All Annotations'') and after these steps (``Filtered'').

\begin{table}[t!]
\centering
\resizebox{\linewidth}{!}{
\footnotesize{
\begin{tabular}{l|ccccccc}
\toprule
      & \cellcolor{lavenderblue}{TBBT} &\cellcolor{lavenderblue}{Buffy} &  \cellcolor{lavenderblue}{Sherlock}  & \cellcolor{lavenderblue}{Friends}&\cellcolor{lavenderblue}{HF}&\cellcolor{lavenderblue}{ALN} &\cellcolor{mistyrose}{Total} \\ \midrule
  All Annotations  & 2,035  & 4,339  &  4,025  &  11,321 & 2,060 & 2,036 & \textbf{27,163} \\ 
    Filtered    & 1,047  & 1,835  &  1,615  &  3,961& 404 & 303 & \textbf{9,165} \\ 
   
\bottomrule
\end{tabular}
}}

\captionof{table}{\small{\textbf{Voice-Track statistics in \dnameemphasis.} The number of voice-tracks for each program set in \dnameemphasis both before and after a filtering step (Section~\ref{sup:sub:data_collection}). All Annotations -- the total voice-track annotations provided with \dnameemphasis. Filtered -- the total voice-track annotations used by our person-clustering method, after ignoring short and overlapping tracks (same as Table 1 in main manuscript). Total -- the summation over all six program sets.}}
\label{tab:voice_tracks}
\end{table}

\section{Metrics}
\label{sup:metrics}
As mentioned in Section 5 in the main manuscript, for each dataset in \dnameemphasis, we use Weighted Cluster Purity (WCP) and Normalized Mutual Information (NMI). 
Furthermore, 
we introduce the metrics of Character Precision and Recall.  
Here, we describe in more detail the WCP and NMI metrics and give some motivation behind the proposed Character Precision and Recall (CP, CR).

\paragraph{Weighted Clustering Purity (WCP).}
WCP weights the purity of a cluster by the number of samples belonging in it; to compute purity, each cluster $c$ containing $n_c$ elements is assigned to the class which is most frequent in the cluster. WCP is highest at 1 when within each cluster, all samples are from the same class. For a given clustering, $\mathcal{C}$, with $\mathcal{N}$ total tracks in the video: $WCP = \frac{1}{N} \sum_{c \in C}^{}  n_c \cdot purity_c $.

\paragraph{Normalized Mutual Information (NMI)~\cite{manning2008introduction}.}
NMI measures the trade-off between clustering quality and number of resulting clusters. 
Given class labels $Y$ and cluster labels $C$, $\text{NMI}(Y,C) = 2\frac{I \left( Y ; C\right)}{ H(Y) + H(C)}$, where $H(.)$ is the entropy and $I(Y;C)=H(Y) - H(Y \backslash C)$ the mutual information.

\paragraph{Character Precision and Recall (CP, CR).}
We introduce Character Precision (CP) and Recall (CR), two metrics computed using the ground truth number of clusters. 
CP is the proportion of tracks in a cluster that belong to its assigned character, 
while CR is the proportion of that character's total tracks that appear in the cluster. 
The assignment is done using the Hungarian algorithm~\cite{kuhn1955hungarian} by using CR as the cost function. 
Note that this assignment is unique, \ie two characters cannot be assigned to the same cluster. 
We measure CP and CR and report results averaged across all characters. 
Our motivation is that the standard metrics are weighted according the number of samples in each cluster, thus disproportionately favouring frequently appearing characters and disregarding tail distributions. 
Instead, similar to character AP~\cite{Nagrani17b}, CP and CR weight all characters equally. 
Similar to the Hungarian matching accuracy used in ~\cite{asano2020labelling,van2020learning}, CP and CR are computed using the ground truth number of clusters. 
Thus, they measure complementary information to WCP and NMI, which do not have access to this information.


\section{Modality Analysis}
\label{sup:modality_analysis}
In this section, we provide further analysis into the discriminative capabilities of each of the three modalities used in \mnameMM (face, body and voice). In Stage 1 of \mnameMM, high-precision clusters are created using just the face modality, as it is the most discriminative of the three. Here, we justify this by instead using the other modalities in Stage 1. Table~\ref{tab:modality_analysis} shows results averaged across all program sets in \dnameemphasis for both AT and OC protocol, when each of the available modalities are used in Stage 1 (termed $\mnameMM_{body}$, $\mnameMM_{voice}$; and $\mnameMM_{face}$). Next, we explain some experimental details, and then analyse these results.
\begin{table}[t!]
\centering

\bgroup
\def\arraystretch{1.1} 
{\setlength\tabcolsep{1.1pt} 
\resizebox{0.7\linewidth}{!}{
\footnotesize{
\begin{tabular}{l|ccc|c|cccc }
\toprule
\multirow{2}{*}{} &  \multicolumn{3}{c|}{\multirow{2}{*}{\textbf{Modality}}}   & \multirow{2}{*}{\textbf{Protocol}} &
\multicolumn{4}{c}{\cellcolor{lavenderblue}{}  }        
 \\

& & &  & &\multicolumn{4}{c}{\multirow{-2}{*}{\cellcolor{lavenderblue}{\textbf{Average}}}}   
\\ \midrule

&  F & B & V    & &
WCP   & NMI                  & CP                 & \multicolumn{1}{c}{CR} \\ 
\hline 
$\mnameMM_{body}$  &  & {\checkmark} &  & AT &       60.6     &      46.9         &     63.4   &  48.1  \\
$\mnameMM_{voice}$ &   & &  {\checkmark} & AT  &      71.0        &  67.9           &  54.6            & \multicolumn{1}{c}{50.3}   
\\ 
$\mnameMM_{face}$  &   {\checkmark} &  &  &  AT &  \textbf{93.4}  &     \textbf{89.4}     &    \textbf{93.0}         & \multicolumn{1}{c}{\textbf{90.2}}  \\ \midrule
$\mnameMM_{body}$  &  &{\checkmark}  &  &OC  &      58.1      &   43.7            &   50.6     & 44.8  \\
$\mnameMM_{voice}$ &   & & {\checkmark} &  OC &      77.5        &    70.4        &   58.1           &  \multicolumn{1}{c}{55.2}   
\\ 
$\mnameMM_{face}$  &   {\checkmark} &  &  & OC &  \textbf{91.7}  &    \textbf{87.2}      &    \textbf{84.7}         & \multicolumn{1}{c}{\textbf{81.9}}  \\ \bottomrule

\end{tabular}
}
}
}
\vspace{-1mm}
\caption{\footnotesize{\textbf{Person-Clustering Results on \dnameemphasis after Stage 1 -- Clustering only speaking person-tracks.} We report the averaged metrics for both AT and OC protocol, averaged across all program sets. Every experiment shown is clustering only a subset of the person-tracks that contain all three modalities (face, body and voice) in order to isolate the clustering performance when each modality is used alone. The three reported methods, $\mnameMM_{body}$, $\mnameMM_{voice}$, $\mnameMM_{face}$, use a different modality as the single modality in Stage 1 (body, voice and face, respectively). The numbers reported are taken after Stage 1.} }
\label{tab:modality_analysis}
\egroup
\end{table}

\begin{figure*}[t!]
\begin{center}
\includegraphics[width=\linewidth]{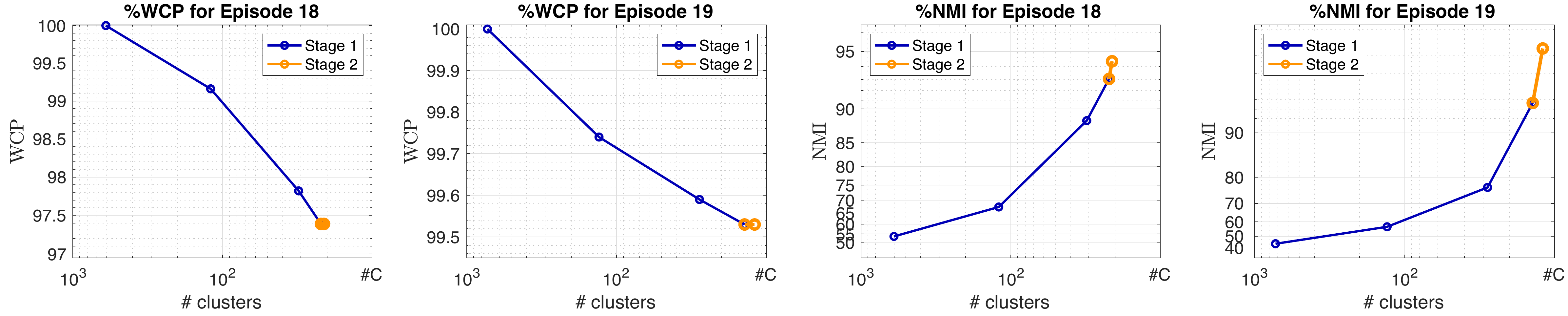} \\
\vspace{3mm}
\includegraphics[width=\linewidth]{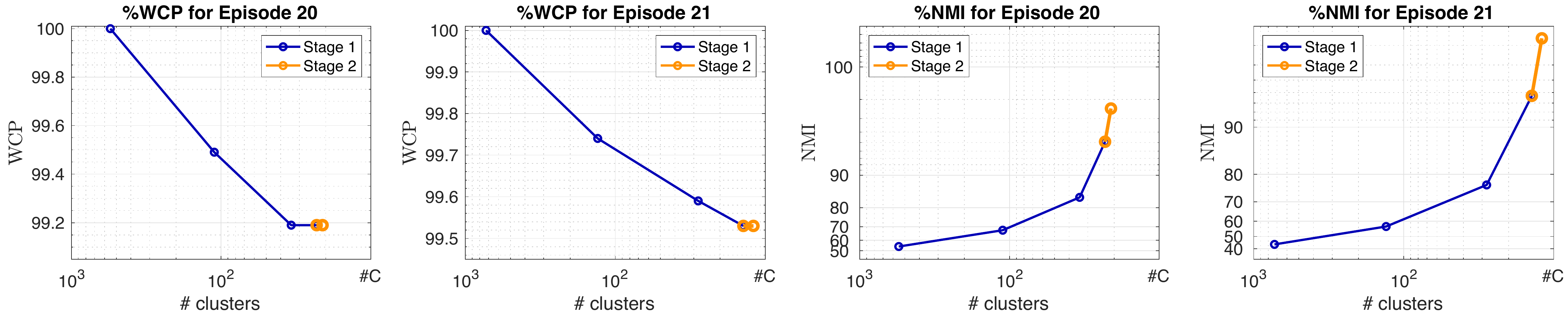} \\
\end{center}
\vspace{-2mm}
  \caption{\small{\textbf{Stage 1 and Stage 2 Person-Clustering results from the program set, Friends.} \%WCP and \%NMI for episodes of Friends from \dnameemphasis, for the Automatic Termination protocol (AT).
  The blue line illustrates the results after Stage 1, while the orange one illustrates the results after Stage 2, \ie bridging clusters by exploiting the voice modality. $\#C$ is the ground truth number of clusters for each episode.
   }}
\label{fig:friends_stage_analysis}
\end{figure*} 
\vspace{2mm}
For fair comparison between $\mnameMM_{body}$, $\mnameMM_{voice}$; and $\mnameMM_{face}$, we cluster the same person-tracks in each of the experiments. This limits the experiments to person-tracks with all three available modalities \ie talking person-tracks with a visible face. To isolate the role of each of the modalities, we report clustering performance after Stage 1. Similarly to $\tau_f^{\text{tight}}$ in \mnameMM, for these experiments we learn nearest neighbour distance thresholds for each modality on the \dnameemphasis val. set.

As shown in Table~\ref{tab:modality_analysis}, only the face modality can be reliably used in Stage 1 to produce high-precision clusters, as reflected by the high values for WCP in both protocol. This justifies the use of the face modality in Stage 1 of \mnameMM. This is understandable, as different identities can sound the same when expressing similar emotions (\eg anger, sadness), and bodies from different identities can look very similar when wearing similar clothing. According to WCP and NMI, $\mnameMM_{voice}$ produces better clustering performance than $\mnameMM_{body}$, indicating that the voice modality is better at discriminating identity than the body modality.

\section{Person-Clustering Results}
\label{sup:person_cluster_res}

In this section, we provide extensive analysis of the person-clustering results obtained by \mnameMM as well as results for an additional experiment. 
First, we explore the impact of Stages 1 and 2 of \mnameMM on some episodes from the Friends program set in \dnameemphasis (Section~\ref{sup:sub:per_stage}). 
Second, we provide further person-clustering results from \mnameMM on \dnameemphasis using the OC protocol. 
Third, we examine the results when clustering tracks from all program sets in \dnameemphasis, concatenated by their research order of broadcast (Section~\ref{su:sub:concatenated}).

\newpage
\subsection{Per-Stage Analysis} 
\label{sup:sub:per_stage}

We examine the effects of Stages 1 and~2 (Section 3 in the main manuscript) on the performance of \mnameMM on episodes from the Friends program set in \dnameemphasis.  
To this end, we plot in Figure~\ref{fig:friends_stage_analysis} the \%WCP and \%NMI results over the number of clusters after each partition of the method for four episodes. 
Each circle in the plot displays the partition (\ie showing the number of clusters of the resulting partition and the corresponding metric value). 
The \textcolor{blue}{blue} lines and circles represent the clustering process at Stage 1 of \mnameMM, while the \textcolor{orange}{orange} ones display the Stage 2 results. 

We observe that in most cases after the first partition (first blue dot) the WCP maintains high values (above 99\%). While Stage 1 progresses, the WCP drops only by a small margin (\ie less than 1\% in most cases), whereas the NMI increases significantly (\ie up to +50\%). This validates that Stage 1 indeed results in high-precision clusters, as the purity (indicated by WCP) is not compromised, and also the NMI increases.

\begin{table*}[ht]
\centering

\bgroup
\def\arraystretch{1.1} 
{\setlength\tabcolsep{1.1pt} 
\resizebox{\linewidth}{!}{
\footnotesize{
\begin{tabular}{lccc|cccc |cccc |cccc |cccc |cccc |cccc |cccc}
\toprule
\multirow{2}{*}{\#} &  \multicolumn{3}{c|}{\multirow{2}{*}{Modality}}   &
\multicolumn{3}{c}{\cellcolor{lavenderblue}{}  }       &\cellcolor{lavenderblue}{\textbf{$\#C_s${=}}}   
& \multicolumn{3}{c}{\cellcolor{lavenderblue}{ }}        &          \cellcolor{lavenderblue}{\textbf{$\#C_s${=}}}                                                 &
\multicolumn{3}{c}{\multirow{2}{*}{\cellcolor{lavenderblue}{}}}  &  \cellcolor{lavenderblue}{\textbf{$\#C_s${=}}} & 
\multicolumn{3}{c}{\multirow{2}{*}{\cellcolor{lavenderblue}{}}} & \cellcolor{lavenderblue}{\textbf{$\#C_s${=}}} & 
\multicolumn{3}{c}{\multirow{2}{*}{\cellcolor{lavenderblue}{}}} & \cellcolor{lavenderblue}{\textbf{$\#C_s${=}}} & 
\multicolumn{3}{c}{\multirow{2}{*}{\cellcolor{lavenderblue}{}}} & \cellcolor{lavenderblue}{\textbf{$\#C_s${=}}} & 
\multicolumn{3}{c}{\multirow{2}{*}{\cellcolor{lavenderblue}{}}} & \cellcolor{lavenderblue}{\textbf{$\#C_s${=}}} \\

& & &  &\multicolumn{3}{c}{\multirow{-2}{*}{\cellcolor{lavenderblue}{\textbf{TBBT}}}}  & \cellcolor{lavenderblue}{130}
& \multicolumn{3}{c}{\multirow{-2}{*}{\cellcolor{lavenderblue}{\textbf{Buffy}} }}        &         \cellcolor{lavenderblue}{165}                                              &
\multicolumn{3}{c}{\multirow{-2}{*}{\cellcolor{lavenderblue}{\textbf{Sherlock}}}}  &  \cellcolor{lavenderblue}{50} & 
\multicolumn{3}{c}{\multirow{-2}{*}{\cellcolor{lavenderblue}{\textbf{Friends}}}} & \cellcolor{lavenderblue}{239} 
&  \multicolumn{3}{c}{\multirow{-2}{*}{\cellcolor{lavenderblue}{\textbf{Hidden Figures}}}}   & \cellcolor{lavenderblue}{10}
&  \multicolumn{3}{c}{\multirow{-2}{*}{\cellcolor{lavenderblue}{\textbf{About Last Night}}}}   & \cellcolor{lavenderblue}{24}
&  \multicolumn{3}{c}{\multirow{-2}{*}{\cellcolor{lavenderblue}{\textbf{Average}}}}   & \cellcolor{lavenderblue}{618} 
\\ \midrule

&  F & B & V    & 
WCP   & NMI                  & CP                 & \multicolumn{1}{c|}{CR} &  
WCP                  & NMI                  & CP                 & \multicolumn{1}{c|}{CR} &  
WCP                  & NMI                  & CP                 & \multicolumn{1}{c|}{CR} & 
WCP                  & NMI                  & CP                 & \multicolumn{1}{c|}{CR} & 
WCP                  & NMI                  & CP                 & \multicolumn{1}{c|}{CR} & 
WCP                  & NMI                  & CP                 & \multicolumn{1}{c|}{CR} & 
WCP                  & NMI                  & CP                 & CR \\ 
\hline 
B-ReID  &  & {\checkmark} &  &  66.2             &       54.9               &       11.4               & 33.0 &           52.0             &       48.8               &       40.6               & 24.6  &          60.3             &          24.0            &        10.7              & 36.0   &         62.8              &           56.0           &       49.8               & 56.4  &  33.7          &           10.5          &           17.6           & \multicolumn{1}{c|}{31.7}    &                27.3      &       17.9               &       19.9               & \multicolumn{1}{c|}{19.3}    &        50.4 & 35.4 & 25.0 & 33.5   \\
B-C1C &  {\checkmark} & {\checkmark} & &       91.7          &       79.1               &       54.4              & \multicolumn{1}{c|}{55.9}    &    74.5               &     62.7                                & \multicolumn{1}{c}{46.8}    &         44.7         &                       77.1               & \multicolumn{1}{c}{44.4}    &   33.2                 &          43.3          &                      88.0  &   82.4 & 74.5 & 78.9  & 
69.5              &     51.8               &               29.7       & \multicolumn{1}{c|}{46.4}    &                  73.1             &     64.8              & \multicolumn{1}{c}{55.7}    &              53.8            &                        79.0 & 64.2 & 49.1 & 53.8 
\\ \midrule
\mnameMM-  &   {\checkmark} &  &  &   94.3     &      85.8             &   84.1            & \multicolumn{1}{c|}{81.8}  &     81.1            &     68.0           &     76.2       & \multicolumn{1}{c|}{76.3}  &       86.79         &     56.87      &    74.8         & \textbf{69.3} &   90.0                & 76.6                 &  90.8              & 82.8 &   \textbf{85.7}     &  \textbf{77.3}          & \textbf{76.7}               &\textbf{ 56.7} &     \textbf{97.9}            &       91.4          &     98.9      &      \multicolumn{1}{c|}{86.9}    &   89.3    & 76.0 & 83.6 &   75.6    \\ 


$\mnameMM_v$ & {\checkmark} &    &  {\checkmark} &    94.3           &        85.8          &             84.1        & \multicolumn{1}{c|}{81.8}    &  81.1 &  68.5      &     76.2          &            75.8      &   86.1   &        62.3        &           72.8    & \multicolumn{1}{c|}{68.8}     &    89.8               &  77.3           &     89.6     & 84.6   & 85.7
             & 77.3             &   76.7   & \multicolumn{1}{c|}{56.7}   &       97.8           &    \textbf{91.9}            &   98.9              & \multicolumn{1}{c|}{87.0}  &    89.3      & 76.4 & 83.4 & 75.5             
\\
$\mnameMM_b$  & {\checkmark} &{\checkmark}  &  &  \textbf{97.7} &   \textbf{93.9}      &    \textbf{86.9}  &       \textbf{83.8}           &     86.9                    &    76.9           &            \textbf{80.0}      & \textbf{79.1}    &     \textbf{87.1}        &           57.5       &    \textbf{74.9}              &  66.8  &    \textbf{94.2}          &     84.6        &  \textbf{95.7}           & 86.0 & 85.6
          &     77.1            &  76.6                & 56.7 &    97.8        &      91.2        &    98.9         &  86.9   &\textbf{91.5}        & 80.2 & \textbf{86.0} & 77.0
\\  \hline 
\mnameMM &  {\checkmark} &{\checkmark}   &{\checkmark}   &    97.7            &      93.9  &     86.9      &   83.8  &     \textbf{86.9}          &              \textbf{77.6}      &      79.8            & 78.5 &      86.4    &        \textbf{63.0}         & 73.1             & 68.7 &     94.0           & \textbf{85.5}                & 94.6            & \textbf{88.1}  &  85.6
          &        77.1       &    76.6        & 56.7  &    97.8        &        91.6          &       \textbf{98.9}            & \textbf{87.0} &    \cellcolor{mistyrose}{91.4}          & \cellcolor{mistyrose}{\textbf{81.5}} & \cellcolor{mistyrose}{85.0} & \cellcolor{mistyrose}{\textbf{77.1}}           
 \\ \bottomrule
\end{tabular}
}
}
}
\vspace{-1mm}
\caption{\footnotesize{\textbf{Person-Clustering Results on \dnameemphasis.} For each program set, each metric is averaged across all episodes. OC protocol. The `Average' column reports averaged metrics across all six program sets. $\#C_s$ is the sum of ground truth clusters across each episode. We report two strong baselines (B-ReID, B-C1C, Section 5.1 in main manuscript) and an ablation on the modalities used. Keys: F-face, B-body, V-voice. \textit{Modality}: used modalities.} }
\label{tab:person_results}
\egroup
\end{table*}

\begin{figure*}[ht!]
\begin{center}
\includegraphics[width=0.98\linewidth]{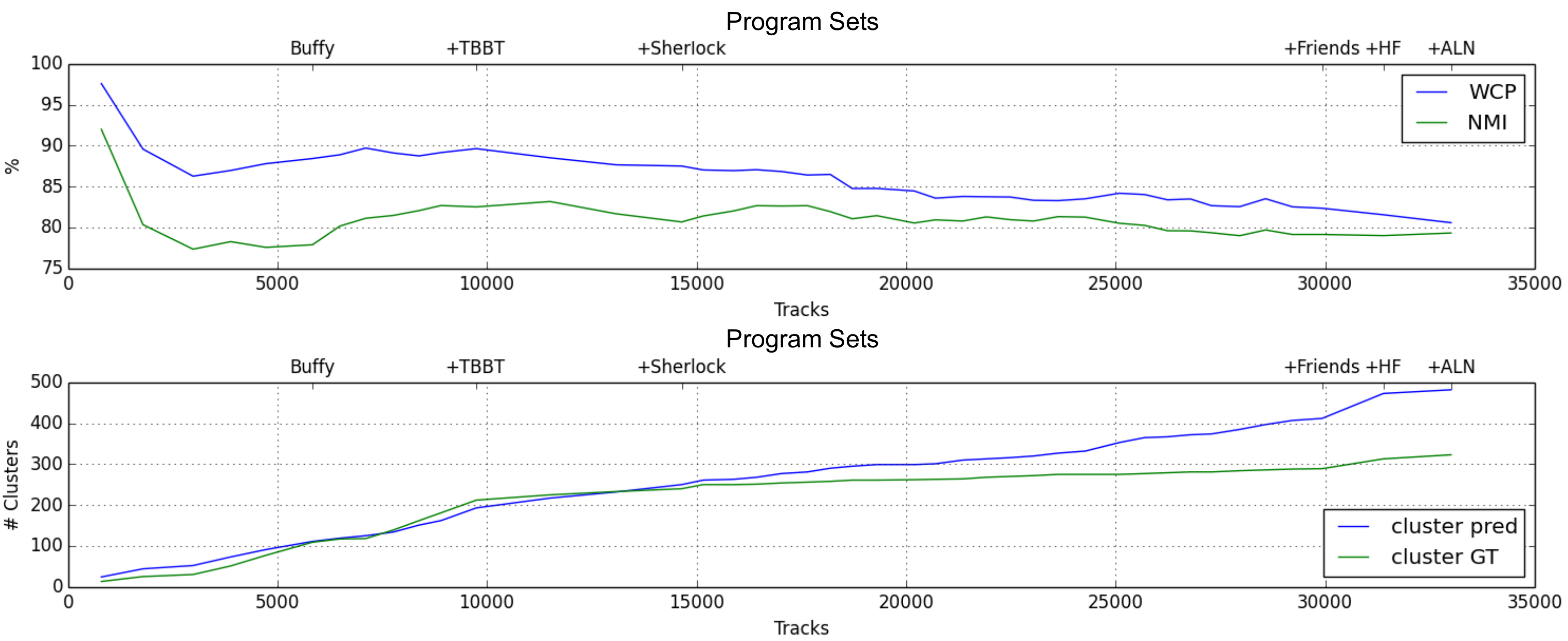}
\end{center}
\vspace{-5mm}
  \caption{\footnotesize{\textbf{Person-Clustering Results when clustering multiple program sets simultaneously.} Incrementally, more and more tracks are considered by adding different program sets together. There are discrete data points for each time the tracks from an additional episode or movie are added. Each data point considers the total cumulative number of tracks up to that point. All experiments are for the Automatic Termination (AT) protocol for person-clustering for \mnameMM. Top: The WCP and NMI measurements. Bottom: The total predicted number of clusters (cluster pred), measured against the ground truth number of clusters (cluster GT). Note that ``cluster GT'' is different to $\#C_s$ in the main manuscript. $\#C_s$ is the summed number of ground truth clusters (number of characters) across multiple episodes. For example, episodes 1 and 2 of Sherlock have 13 and 22 ground truth clusters, respectively. In this case, $\#C_s = 35$. However, some characters appear in both episodes, such as ``John'' or ``Sherlock''. Instead, ``cluster GT'' is the total number of \emph{unique} ground truth characters and therefore clusters across multiple episodes. For the same example of episodes 1 and 2 of Sherlock, ``cluster GT'' is equal to 31, as 4 characters feature in both episodes.
  }
  }
  \vspace{0mm}
\label{fig:cluster_sim}
\end{figure*}

The orange dots signify the additional partition from Stage 2. Stage 2 consistently and significantly increases the NMI of the resulting clusters (\ie by up to 5\%), without sacrificing their purity (WCP remains constant). This indicates that Stage 2 bridges high-precision clusters of the same identity, thus retaining the high WCP, while decreasing the identity overlap between clusters. 


\newpage

\subsection{Oracle Clusters Results}

\label{sup:sub:at}

Table~\ref{tab:person_results} gives person-clustering results for the OC protocol. The experiments, ablation studies and baselines are the same as those used for the AT protocol, and explained in Section 5.1 of the main manuscript. Similarly to the AT protocol, \mnameMM-- significantly outperforms both baselines across all metrics and program sets. \mnameMM gives a further boost when averaged across all program sets. The voice modality provides comparably less of a performance boost in the OC protocol (here) relative to the AT protocol (Table 2 in the main manuscript). This is due to the Oracle Cluster protocol (OC), which forces the clusters to merge beyond the automatic termination point until the ground truth number of clusters is reached. Next, we explain this in further detail.

\mnameMM automatically stops clustering when the features within each cluster can no longer confidently be used to discriminate between clusters of the same identity. To reach the oracle number of clusters, the clusters are merged in a non-discriminative way. In this case, this reverses the positive impact of the voice modality (seen in Table 2 in the main manuscript) by merging the new clusters incorrectly until the oracle number of clusters is reached. This opens possibilities for future research into more effective ways of reducing to the ground truth number of clusters. The Automatic Termination protocol is the more realistic setting for real-world deployment of person-clustering algorithms on videos with unknown numbers of characters.

\subsection{Clustering on Multiple Program Sets Simultaneously}
\label{su:sub:concatenated}
In this section, we present results for the person-clustering task when clustering tracks from multiple program sets simultaneously. In the main manuscript, all experiments are conducted on individual program sets from  \dnameemphasis. Here, we cluster tracks from multiple program sets at the same time. In detail we incrementally consider additional episodes and movies from each of the program sets. Results for the WCP, NMI and the number of predicted clusters against the ground truth number of characters for the AT protocol for person-clustering are shown in Figure~\ref{fig:cluster_sim}. The order with which program sets are added to the clustering experiment is in line with the timing of their first use in Computer Vision research (\ie first Buffy~\cite{Everingham06a}, followed by TBBT~\cite{TVD}, then Sherlock~\cite{Nagrani17b} and so on). Episodes within each of the TV-shows are added chronologically (starting with the first episode in the program set).

Impressively, Figure~\ref{fig:cluster_sim} shows that when clustering all tracks from \dnameemphasis simultaneously, the WCP and NMI remain high at 80.6\% and 79.3\%, respectively. 
This indicates that most clusters have high purity, even with 323 different characters and over 30,000 tracks, over the visually disparate TV-shows and movies. 
As expected, these metrics drop as the total number of tracks increases, as the task becomes much more difficult. 
Until the introduction of tracks from episodes from Friends (14,642 tracks), the predicted number of clusters lies very close to the ground truth number of clusters. This indicates that \dnameemphasis is accurately predicting the number of different characters in the tracks. 
As the total number of tracks increases, the predicted number of clusters diverges from the ground truth number, and \mnameMM predicts more clusters than there are characters. 
This is in line with and partially explained by the combination of cannot-link constraints and decreasing WCP. As the purity of clusters decreases, the cannot-link constraints start preventing clusters containing tracks of the same identity from merging. This results in \mnameMM automatically terminating the  clustering when there are more clusters than characters. We observe similar results when adding datasets in different orders. Similar experiments for combining the TBBT and Buffy datasets for face-clustering are presented in~\cite{ball}.






\section{Face-Clustering Results}
\label{sup:face_cluster_res}

Here, we give further analysis of the face-clustering results shown in Table 3 of the main manuscript (and repeated in Table~\ref{tab:face_results}). This is an extension of Section 5.3 in the main manuscript. In detail, the extra analysis concerns the automated termination (AT) criterion, and the relation of \mnameMM to previous methods. To summarise Section 5.3 of the main manuscript, \mnameMM significantly surpasses the performance of previous methods across all program sets, all metrics and both AT and OC protocol.

First, we analyse the AT protocol results. The goal of the AT protocol is to automatically terminate clustering and assess the quality of the resulting clusters. This is a realistic protocol for videos in-the-wild where the number of characters is unknown. Here, the number of predicted clusters, $\#C_p$, can be measured relative to the ground truth number of clusters, $\#C_s$. In all program sets, \mnameMM predicts more clusters than the ground truth. This is because \mnameMM prioritises high-precision. For TBBT, $\#C_p$ is very close to $\#C_s$ (168 vs.\ 130), and is in fact closer than the predictions of all previous methods. This is impressive seeing as the goal of BCL~\cite{ball} is to predict the ground truth number of clusters. For the other program sets, $\#C_p$ is slightly further from $\#C_s$ than previous methods (\eg a difference from $\#C_s$ of 36 for Sherlock vs.\  25 for C1C~\cite{Kalogeiton20}). We now give two reasons why despite this, the clusters from \mnameMM are far more desirable than those from previous methods.

\begin{figure*}[htb]
    \centering 
\begin{subfigure}{0.25\textwidth}
  \includegraphics[width=\linewidth]{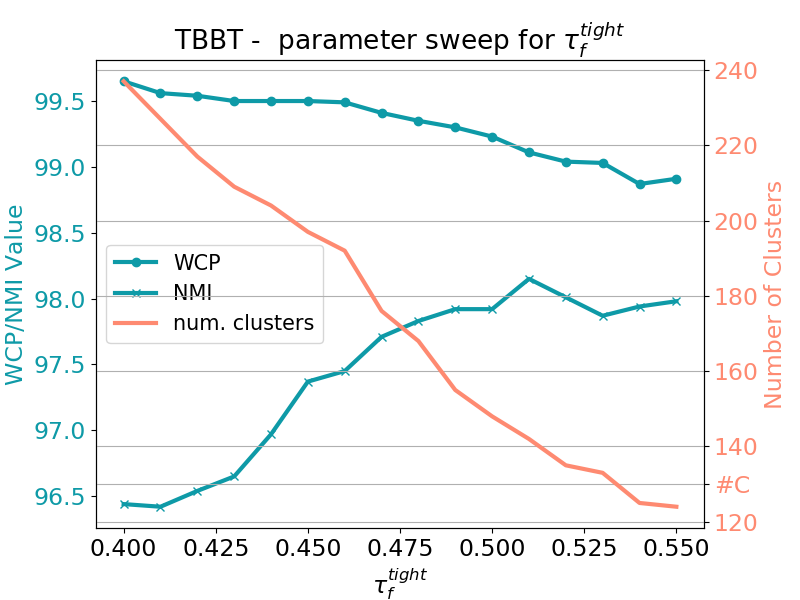}
  \caption{TBBT}
  \label{fig:1}
\end{subfigure}\hfil 
\begin{subfigure}{0.25\textwidth}
  \includegraphics[width=\linewidth]{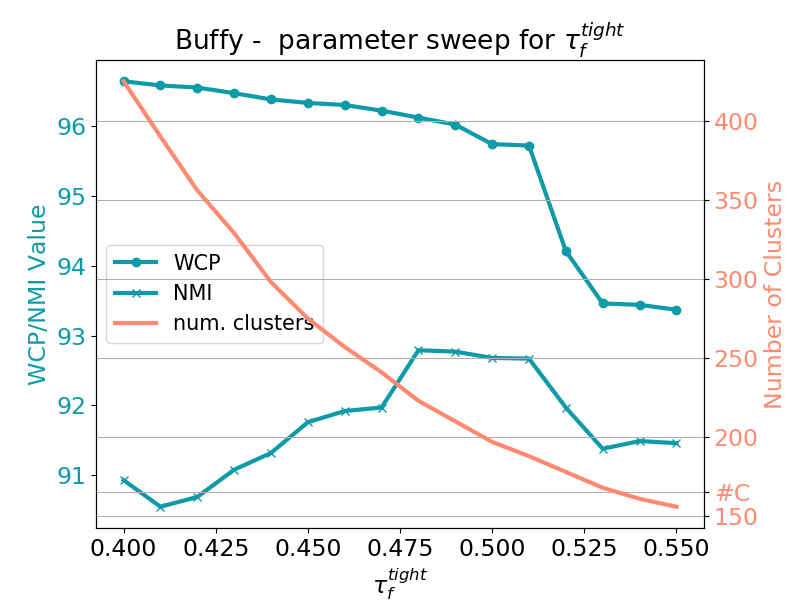}
  \caption{Buffy}
  \label{fig:2}
\end{subfigure}\hfil 
\begin{subfigure}{0.25\textwidth}
  \includegraphics[width=\linewidth]{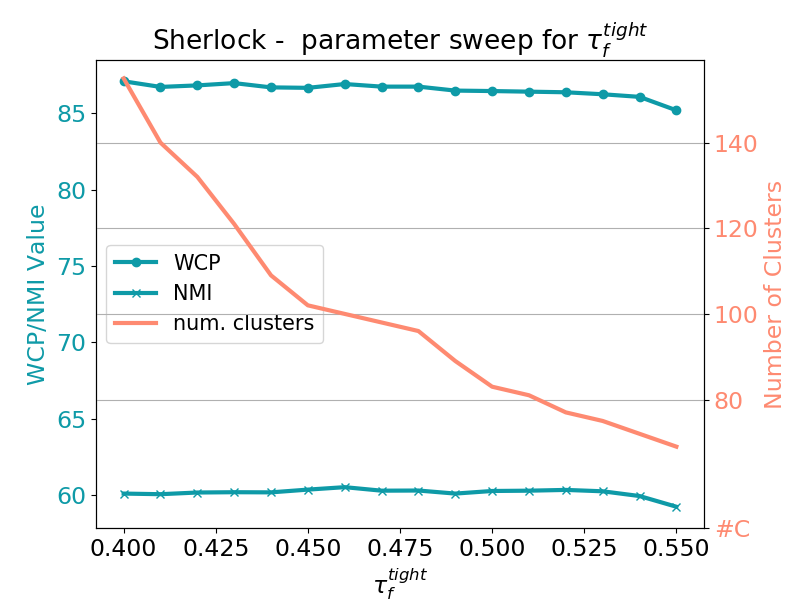}
  \caption{Sherlock}
  \label{fig:3}
\end{subfigure}

\medskip
\begin{subfigure}{0.25\textwidth}
  \includegraphics[width=\linewidth]{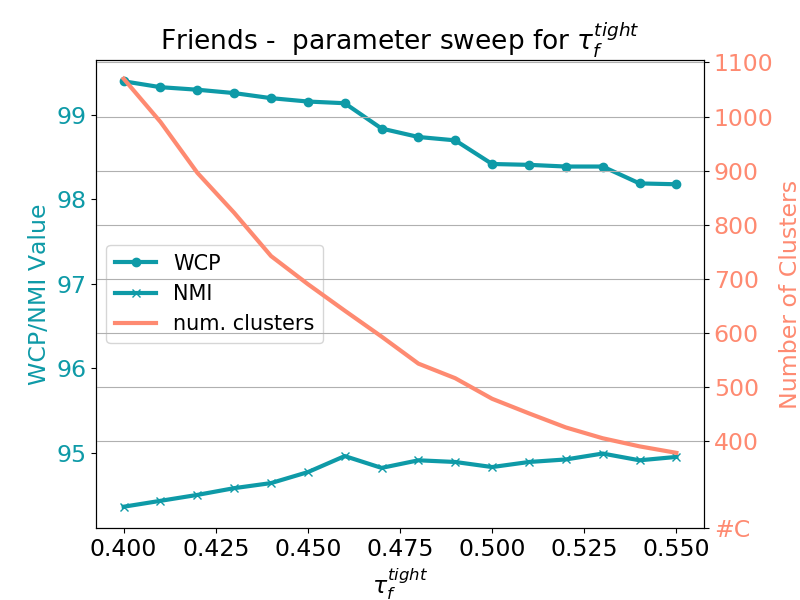}
  \caption{Friends}
  \label{fig:4}
\end{subfigure}\hfil 
\begin{subfigure}{0.25\textwidth}
  \includegraphics[width=\linewidth]{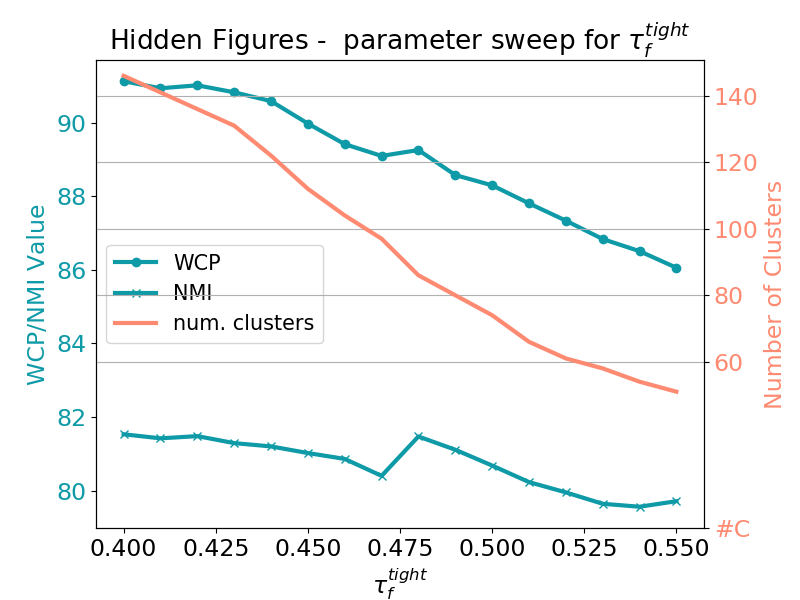}
  \caption{Hidden Figures}
  \label{fig:5}
\end{subfigure}\hfil 
\begin{subfigure}{0.25\textwidth}
  \includegraphics[width=\linewidth]{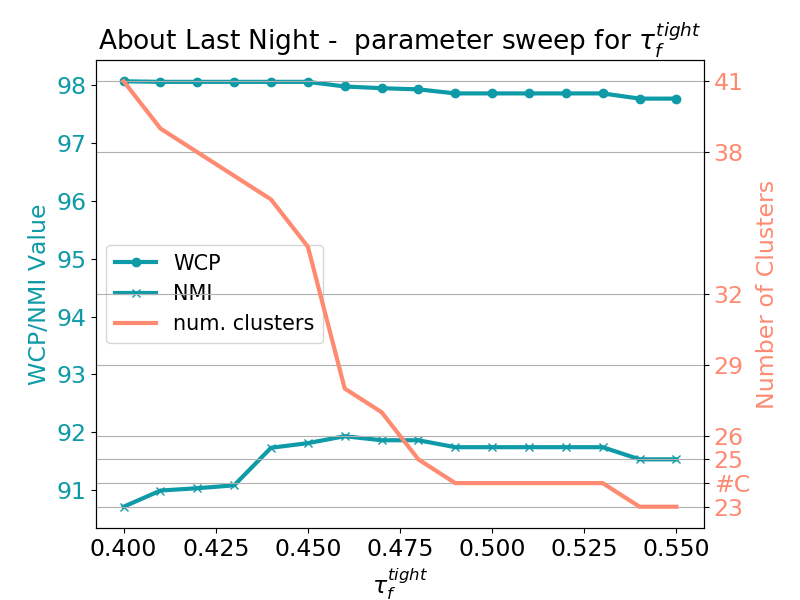}
  \caption{About Last Night}
  \label{fig:6}
\end{subfigure}
\caption{\footnotesize{\textbf{Parameter sweep for $\tau_f^{\text{tight}}$ on the six program sets in \dnameemphasis.} For each program set, the NMI, WCP and number of clusters are plotted, for the Automatic Termination criterion, for varying values of $\tau_f^{\text{tight}}$. We additionally show for each program set, the ground truth number of clusters, \#C, marked on the Number of Clusters axis of each plot. For the numerical values of \#C, we refer the reader to Table 2 in the main manuscript.  }
}
\label{fig:param_Sweeps}
\end{figure*}

First, the clusters from \mnameMM are far higher quality. It would be expected that when predicting more clusters than there are ground truth clusters, any method would achieve higher WCP. However, NMI is also significantly higher for \mnameMM than previous methods (\eg on average 9.8\% higher than the best prior work across all program sets). Second, for downstream applications, it is far more useful to have many high-precision clusters, than few very low-precision clusters. The latter in this case requires a large amount of human labelling in order to correctly label the person-tracks from the clusters (a cluster property reflected by the \textit{Operator Clicks Index} (OCI-k)~\cite{isthatyou} metric). Furthermore, a good way of measuring the utility of clusters for a downstream task is the character precision and recall metrics. These metrics assign each character uniquely to a cluster, and measure the resulting precision and recall of these pseudo-labels. \mnameMM significantly achieves a CP and CR of 56.0\% and 39.3\% higher, respectively, than C1C across all program sets. This indicates that although prior work may predict a number of clusters closer to the ground truth than \mnameMM, these clusters however are of almost no use for downstream applications, unlike the clusters from \mnameMM.

Next, we discuss \mnameMM in relation to previous methods. C1C continues using face to cluster even when there are large distances between clusters, and therefore degenerates in the later partitions, leading to lower WCP and NMI. Unlike BCL,  \mnameMM uses pre-trained features, thus alleviating the computational burden of training, allowing for greater generalisation, and as we demonstrate leading to better results. BCL uses the assumption that each identity occupies the same hyper-spherical volume in their learnt latent space. We argue that complex similarity structures and variation between faces of the same identity mean that they cannot be constrained to within fixed-radius hyper-spheres (BCL), even when training with this objective. Instead, \mnameMM does not enforce such a constraint, and uses a nearest neighbour constraint with multi-modality to connect highly dissimilar tracks.



\section{~Parameter Selection \& Sweeps}
\label{sec:results}

In this section, we give a parameter sweep for the nearest neighbour distance threshold $\tau_f^{\text{tight}}$ (Section 3.1 in main manuscript), and give further description and analysis on the automatic parameter selection method for $\tau_v^{\text{loose}}$ (Section 3.2 in main manuscript).

\subsection{Nearest Neighbor Distance Threshold} 
Here, we give metrics across all program sets in \dnameemphasis for parameter sweeps on the nearest neighbour distance threshold, $\tau_f^{\text{tight}}$. These are displayed in Figure~\ref{fig:param_Sweeps}. As detailed in the main manuscript, the value was chosen on the validation partition of \dnameemphasis. To isolate the role of $\tau_f^{\text{tight}}$, all metrics are evaluated at the Automated Termination criterion, after Stage 1, and using only the face-track annotations. The metrics at the chosen value of $\tau_f^{\text{tight}} = 0.48$, are therefore equivalent to \mnameMM-- at AT protocol in Table 3 in the main manuscript. We notify the reader that in the main manuscript, it reads that $\tau_f^{\text{tight}} = 0.52$. This is incorrect, the value is $\tau_f^{\text{tight}} = 0.48$. 

Across most program sets, the same relationship between the metrics and $\tau_f^{\text{tight}}$ is seen. Namely, as $\tau_f^{\text{tight}}$ increases, NMI increases, while WCP and the total number of clusters decreases. In more detail, as $\tau_f^{\text{tight}}$ increases, the maximum distance at which clusters can merge increases. This leads to more cluster merges before the automatic termination of Stage 1. This is reflected by the decreasing number of clusters at the termination point. Firstly, there is an increased likelihood of incorrect merges, where clusters depicting different identities merge together, leading to lower precision clusters, as shown by decreasing WCP. Increasing $\tau_f^{\text{tight}}$ also leads to more correct merges. This is reflected by the rising NMI, which shows that the identity overlap between clusters is decreasing. An increasing NMI can be interpreted as there being more correct merges than incorrect merges. In some program sets (\eg Buffy, Sherlock), NMI starts to decrease as $\tau_f^{\text{tight}}$ increases, indicating that more incorrect merges are being made than correct merges.

In a window surrounding the learnt value of 0.48, the NMI and WCP are roughly stable at very high values across all program sets (high relative to the respective prior work on those program sets - see Table 3 in main manuscript). This demonstrates that this learnt parameter generalises well to the different program sets, that the face features are indeed universal; and that \mnameMM is not particularly sensitive to this choice of parameter. The program sets in \dnameemphasis are highly visually disparate. These results therefore indicate that \mnameMM could be simply and effectively used on any number of \emph{different program sets} not in \dnameemphasis.

At the chosen value of $\tau_f^{\text{tight}} = 0.48$, often more clusters are predicted than the ground truth number (marked as \#C in Figure~\ref{fig:param_Sweeps}). In some program sets, this is by just a small number (168 vs $\#C = 130$ for TBBT, 223 vs $\#C = 165$ for Buffy). There is a trade-off between obtaining a number of clusters similar to \#C, and the precision of these clusters. Our design choice at Stage 1 is to produce clusters with very high-precision. Stage 2 leads to a further reduction of these clusters by using multiple modalities to merge clusters. A discussion in Section~\ref{sup:face_cluster_res} explains why over-predicting the number of clusters is beneficial for downstream uses of the clusters.

\begin{figure}[t!]
\vspace{-2mm}
     \centering
     \begin{subfigure}[b]{0.23\textwidth}
         \centering
         \includegraphics[width=\textwidth]{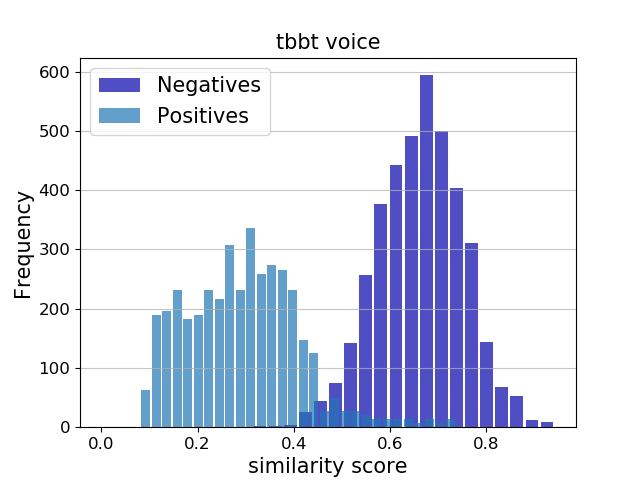}
        \vspace{-1mm}
         \caption{\footnotesize{ TBBT }} 
         \label{SU_GT}
         
     \end{subfigure}
     \hfill
     \begin{subfigure}[b]{0.23\textwidth}
         \centering
         \includegraphics[width=\textwidth]{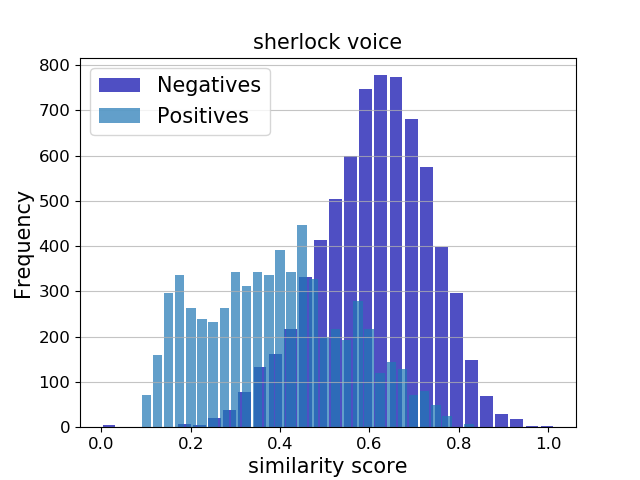}
         \vspace{-1mm}
         \caption{\footnotesize{  Sherlock  }}
         \label{SU_MMHPC_absolute}
     \end{subfigure}
     \vspace{-1mm}
\caption{\footnotesize{\textbf{Voice similarities in two program sets from \dnameemphasis.} Here we show similarities between voices of the same identity (positives) and different identities (negatives). These are found via the cannot-link constraints (negatives) and the clusters from Stage 1 (positives and negatives). Similarities are computed via (1 minus cosine similarity). This process finds less positives than negatives, hence the frequency of the positives is scaled to match that of the negatives. }
}
        \label{fig:voice_sims}
\end{figure}

\begin{table}[t!]
\centering
\resizebox{0.75\linewidth}{!}{
\footnotesize{
\begin{tabular}{l|ccccccc}
\toprule
      & \cellcolor{lavenderblue}{TBBT} &\cellcolor{lavenderblue}{Buffy} &  \cellcolor{lavenderblue}{Sherlock}  & \cellcolor{lavenderblue}{Friends}&\cellcolor{lavenderblue}{HF}&\cellcolor{lavenderblue}{ALN}  \\ \midrule
  $\tau_v^{\text{loose}}$  & 0.36  & 0.17  &  0.19  &  0.31 & 0.19 & 0.33 \\ 

\bottomrule
\end{tabular}
}}
\captionof{table}{\small{\textbf{The automatically learnt values for $\tau_v^{\text{loose}}$ for the different program sets in \dnameemphasis.} }}
\label{tab:voice_loose_table}
\end{table}

\subsection{Automatically Learnt Hyper-Parameters}
The values for the threshold on the voice similarities that are used in the multi-modal bridges, $\tau_v^{\text{loose}}$, are learnt \emph{automatically} for each of the audibly disparate program sets in \dnameemphasis (this is detailed in Section 3.4 in the main manuscript). Here, we give the values that are learnt for each program set, provide some analysis, and visualise the voice distances that the hyper-parameters were learnt from.

The values of $\tau_v^{\text{loose}}$ learnt automatically for the different program sets are given in Table~\ref{tab:voice_loose_table}. 
The voice distances between different identities are found via a combination of cannot-link constraints and the clusters from Stage~1. 
We observe that for some program sets these voice distances are quite high. 
This in turn leads to a relatively high value of $\tau_v^{\text{loose}}$ (\eg TBBT, Friends).
We additionally show the similarities between voices for the same identity (positives) and different identities (negatives) in Figure~\ref{fig:voice_sims} for two program sets from \dnameemphasis.

A high value of $\tau_v^{\text{loose}}$ indicates that the characters all sounded different to the voice embedding network, and in turn the respective features from different speakers were able to be separated in the embedding space (Figure~\ref{fig:voice_sims} - left). For the multi-modal bridges, this means that the voices from two speaking person-tracks can sound quite different and a bridge can still confidently be formed.

For other program sets, the voice distances between the different identities are quite low, and therefore $\tau_v^{\text{loose}}$ is also low (\eg Buffy, Sherlock). In these cases, there are many similar sounding characters; hence, the voice embedding network cannot separate the embeddings from different identities well (Figure~\ref{fig:voice_sims} - right). For the multi-modal bridges, this means that the voices from two speaking person-tracks must sound very similar for a bridge to  still confidently be formed, as only then can the voice modality (together  with the concurrent agreement from the face modality) be sure that it is the same person.


\end{document}